\documentclass{article}

\usepackage[preprint]{neurips_2025}

\usepackage[utf8]{inputenc} % allow utf-8 input
\usepackage[T1]{fontenc}    % use 8-bit T1 fonts
\usepackage{hyperref}       % hyperlinks
\usepackage{url}            % simple URL typesetting
\usepackage{booktabs}       % professional-quality tables
\usepackage{amsfonts}       % blackboard math symbols
\usepackage{nicefrac}       % compact symbols for 1/2, etc.
\usepackage{microtype}      % microtypography
\usepackage{xcolor}         % colors

\usepackage{amsmath}
\usepackage{pifont}
\usepackage{graphicx}
\usepackage[capitalize,noabbrev]{cleveref}
\usepackage{wrapfig}
\usepackage{enumitem}
\usepackage[bottom]{footmisc}
\usepackage{multirow}
\usepackage{bbm}

\usepackage{amsthm}
\usepackage{algorithm}
\usepackage{algorithmic}
\usepackage{bm}
\usepackage[capitalize,noabbrev]{cleveref}
\usepackage{mathtools}
\usepackage{ulem}

\theoremstyle{plain}

\theoremstyle{definition}

\theoremstyle{remark}

\title{RL2Grid: Benchmarking Reinforcement Learning\\in Power Grid Operations}

\author{%
  Enrico Marchesini\\
  Massachusetts Institute of Technology\\
  \texttt{emarche@mit.edu}\\
  \And
  Benjamin Donnot\\
  RTE France\\
  \texttt{benjamin.donnot@rte-france.com}\\
  \And
  Constance Crozier\\
  Georgia Institute of Technology\\
  \And
  Ian Dytham\\
  National Grid ESO\\
  \And
  Christian Merz\\
  50Hertz\\
  \And
  Lars Schewe\\
  University of Edinburgh\\
  \And
  Nico Westerbeck\\
  50Hertz\\
  \And
  Cathy Wu\\
  Massachusetts Institute of Technology\\
  \And
  Antoine Marot\\
  RTE France\\
  \texttt{antoine.marot@rte-france.com}\\
  \And
  Priya L. Donti\\
  Massachusetts Institute of Technology\\
  \texttt{donti@mit.edu} 
}

\begin{document}

\maketitle

\begin{abstract}
Reinforcement learning (RL) can provide adaptive and scalable controllers essential for power grid decarbonization. However, RL methods struggle with power grids' complex dynamics, long-horizon goals, and hard physical constraints. For these reasons, we present RL2Grid, a benchmark designed in collaboration with power system operators to accelerate progress in grid control and foster RL maturity. Built on RTE France's power simulation framework, RL2Grid standardizes tasks, state and action spaces, and reward structures for a systematic evaluation and comparison of RL algorithms. Moreover, we integrate operational heuristics and design safety constraints based on human expertise to ensure alignment with physical requirements. 
By establishing reference performance metrics for classic RL baselines on RL2Grid's tasks, we highlight the need for novel methods capable of handling real systems and discuss future directions for RL-based grid control.\footnote{Code is available at \url{https://github.com/emarche/RL2Grid}}

% old version
%Reinforcement learning (RL) has the potential to transform power grid operations by providing adaptive, scalable controllers essential for grid decarbonization and resilience.
% However, despite their promise, today's RL methods struggle to deal with complex dynamics, aleatoric uncertainty, long-horizon goals, and hard physical constraints, limiting their application in challenging real-world problems. This paper presents RL2Grid, a benchmark developed in collaboration with European power system operators to accelerate progress in grid control and foster the maturity of RL.
% Our work builds upon Grid2Op, a power grid simulation framework developed by RTE France, to provide standardized tasks, state and action spaces, and rewards within a common interface, presenting a common basis for monitoring and promoting progress. Additionally, we formalize heuristic-guided training and safety constraints derived from human operator knowledge and safe practices to reflect grid requirements. We evaluate and compare widely adopted RL algorithms across the increasingly complex settings represented within RL2Grid, establishing reference performance metrics and offering insights into the effectiveness of different approaches. Our findings indicate that power grids present substantial challenges for modern RL, underscoring the need for novel methods capable of dealing with complex real-world physical systems.
\end{abstract}

%%%%%%%%%%%%%%%%%%%%%%%%%%%%%%%%%%%%%%%%%%%%%%%%%%%%%%%%%%%%%%%%
%% Section: Submission of papers to RLJ/RLC
%%%%%%%%%%%%%%%%%%%%%%%%%%%%%%%%%%%%%%%%%%%%%%%%%%%%%%%%%%%%%%%%
\section{Introduction}
\label{sec:introduction}

\begin{figure}[t]
    \centering
    \vspace{-5pt}

    \includegraphics[width=\textwidth]{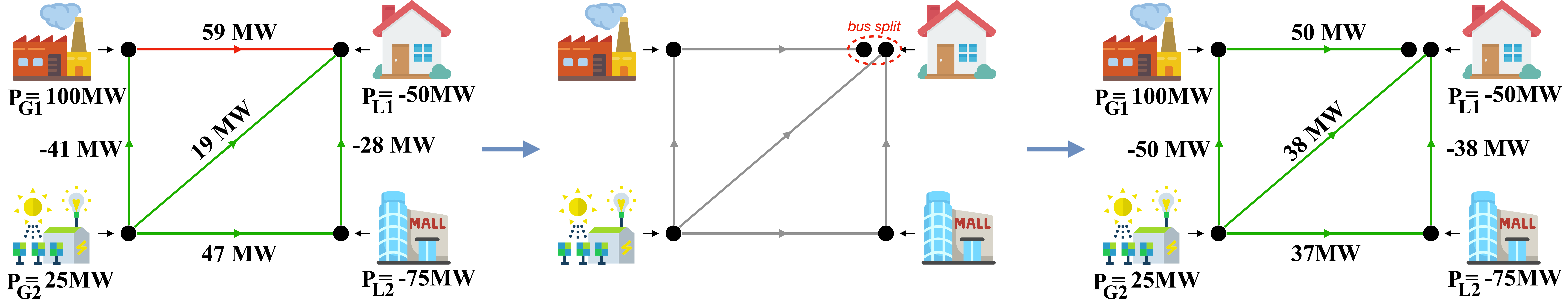}
    \caption{A high-level overview of a power grid action (bus split) to address an overloaded line (red).}
    \label{fig:power_grid_example}
    \vspace{-10pt}
\end{figure}

Power grids require a rapid transition to low-carbon energy and improved robustness against climate-induced extremes in order to combat climate change.
This requires operating under increasing speed, scale, and uncertainty, due in large part to evolving supply and demand profiles resulting from distributed devices and variable renewable energy sources (VREs) \citep{rlgrid_survey}. This integration creates significant challenges for human operators and traditional power system solvers \citep{marot2022perspectives}. To clarify what a power grid is, \Cref{fig:power_grid_example} shows a simplified scenario with four substations (dots) interconnected by transmission lines (edges), two power generators, and two loads connected to buses within each substation. Generators produce power that flows through transmission lines to meet demands (loads). Transmission leads to power losses due to resistive heat on the lines, and substations (which may contain multiple buses) can act as ``switches'' to direct power flows to an extent. All these electrical components have physical constraints that must be satisfied (e.g., generators have ramping limits preventing arbitrary instantaneous changes in power output, and transmission lines have maximum capacities, with prolonged overloads causing disconnections and permanent damage). 

Deep reinforcement learning (RL) is a promising approach for power grid operations, having demonstrated impressive control performance over the last decade \citep{dqn, alphago, gtsophy}. 
However, \textit{power grids encompass many open research questions in RL}, including dealing with complex dynamics, aleatoric uncertainty, learning long-horizon goals, and satisfying hard physical constraints. Investigating realistic power grid tasks from an RL perspective could thus yield substantial benefits for both society and the RL research community. Nonetheless, progress in relevant RL methodologies is hindered by a lack of standardized benchmarks that can help promote and monitor progress, identify bottlenecks, and develop insights to address real-world challenges. 
We fill this gap by introducing RL2Grid, an RL benchmark for realistic power grid operations designed in collaboration with major transmission system operators (TSOs).
% ---RTE France, 50 Hertz, National Energy System Operator (NESO). 
RL2Grid aims to accelerate progress in grid control and advance RL methods tailored for real-world problems by modeling a diverse, standardized set of increasingly complex power grid environments for RL research. %that involve dealing with the combinatorially large number of possible actions available in typical grid operations. 
These tasks build upon RTE France's Grid2Op \citep{grid2op}, a realistic power grid simulation framework, and are presented within a standard Gymnasium-based interface, alongside common state and action spaces, and rewards, to provide a shared base for comparison. We additionally perform an in-depth analysis of RL2Grid's design choices by investigating the quality of the actions available in different task settings. To incorporate power grids' real operation practices and hard physical requirements, %(e.g., satisfying power flows, generator and line limits, as well as avoiding islanding)
we also introduce a heuristic module incorporating common line reconnection and idle practices in the grid dynamics \citep{grid2op}, as well as constrained task formalizations for safe RL \citep{gu2024review}. Finally, by extending the well-known CleanRL library \citep{huang2022cleanrl} to include flexible configurations for algorithm implementation details, we conduct a comprehensive empirical comparison of classic RL algorithms that are frequently used in the literature as baselines or building blocks for more complex approaches. Our codebase is available as supplementary material.

% We also present a heuristic module facilitating the seamless incorporation of basic grid operations (e.g., line reconnection and idle actions) into the training loop of existing algorithms, which we confirm yields a drastic improvement in performance and sample efficiency across all RL algorithms.
%; and (ii) a constrained formalization to describe the safety specifications commonly encountered in power systems (e.g., satisfying power flows, generator limits, and line limits, as well as avoiding islanding). 
%These tasks build upon RTE's Grid2Op \citep{grid2op}, a well-regarded simulation framework for realistic power grid operations, and are presented within a standard Gymnasium-based interface, alongside common rewards, state spaces, and action spaces to provide a common basis for comparison.

Finally, our collaboration with TSOs allows us to extensively discuss (\Cref{sec:discussion}): (i) directions to further improve the realism of power grid simulators to enable last-mile development and deployment of the methodological advances we hope RL2Grid will promote; and (ii) the relationship of power grid operations to open challenges in RL. Through RL2Grid, we aim to foster the maturity of RL methods for real-world power grid domains and provide a standardized basis for comparative analysis.% by providing realistic tasks that encompass important open questions, and a standardized basis for comparative evaluation and analysis. 

\section{Preliminaries}
\label{sec:preliminaries}

%\subsection{Reinforcement Learning Formalization}

A power grid task can be modeled as a Markov decision process (MDP)---a tuple $(\mathcal{S}, \mathcal{A}, \mathcal{P}, \rho, R, \gamma)$, where $\mathcal{S}$ and $\mathcal{A}$ are the finite sets of states and actions, respectively, $\mathcal{P}:\mathcal{S \times \mathcal{A} \times \mathcal{S}} \to [0, 1]$ is the state transition probability distribution, $\rho: \mathcal{S}\to [0, 1]$ is the initial uniform state distribution, $R: \mathcal{S}\times\mathcal{A} \to \mathbb{R}$ is a reward function, and $\gamma\in [0,1)$ is the discount factor. In policy optimization algorithms, agents learn a parameterized policy $\pi:\mathcal{S} \times \mathcal{A} \to [0,1]$, modeling the probability of taking an action $a_t \in \mathcal{A}$ in a state $s_t \in \mathcal{S}$ at a certain step $t$. In value-based algorithms, agents learn state and/or action value functions $V_\pi$ and $Q_\pi$% (Equation \ref{eq:value_functions})
, representing the expected discounted return when starting from a state $s$ (and action $a$ for $Q_\pi$) and following the policy $\pi$ thereafter. In these contexts, agents typically use a greedy policy taking the action corresponding to argmax over $Q_\pi$. The goal is to find a policy that maximizes the expected discounted return $\mathbb{E}_\pi [\sum_{t=0}^\infty \gamma^t R(s_t, a_t)]$.%\footnote{For the sake of clarity and brevity, we refer to \citet{grid_mdp} for an exhaustive overview of the MDP formalization, the state and action spaces, reward, and value ranges mentioned in the next sections.} 

% \begin{equation}
%     V_\pi(s) = \mathbb{E}_\pi\Bigg[ \sum_{t=0}^\infty \gamma^t R(s_{t}, a_{t}) \vert s_0 = s\Bigg],~\hfill Q_\pi(s, a) = \mathbb{E}_\pi\Bigg[ \sum_{t=0}^\infty \gamma^t R(s_{t}, a_{t}) \vert s_0 = s, a_0 = a\Bigg].
% \label{eq:value_functions}
% \end{equation}

%Given the current state and action, we can also measure how much better or worse the agent performs compared to its expected performance using the advantage function $A_\pi(s,a) = Q_\pi(s,a) - V_\pi(s)$. In these contexts, agents typically use a greedy policy over the action value or the advantage function (i.e., they take the action corresponding to argmax over the values). 
To promote safety, we also model grid tasks as constrained MDPs (CMDPs) \citep{LagrangianMult1}, %---the most common formalism used in safe RL today \citep{gu2024review}. 
by adding a set of constraints $\mathcal{C} := \{C_i\}_{i=1,\ldots,n}$ defined over unsafe state and action pairs identified by indicator cost functions. %(a positive value deems a pair as unsafe). 
These can be used to describe both instantaneous constraints which must be satisfied at every point in time, and cumulative constraints specifying limits on the accumulation of cost over a specified horizon. % Although the instantaneous case better describes the safety specifications of common power systems (e.g., satisfying power flows and generator limits), cumulative constraints are often easier to accommodate within existing algorithms \citep{su2024review}. 
For instance, policy optimization approaches typically transform the CMDP into an equivalent unconstrained Lagrangian optimization problem $\mathcal{L}$ over the policy parameters using dual variables as $\mathcal{L}^{\pi}(\bm{\lambda}) = J_R + \mathcal{L}_\mathcal{C}(\bm{\lambda})$, where $\mathcal{L}_\mathcal{C}(\bm{\lambda}) = -\sum_{i=1}^{n}\lambda_i(V^{\pi}_{C_i} - \tau_i)$, $J_R$ is the return objective to maximize, $\bm{\lambda} = \{\lambda_i\}_{i=1,\dots,n}$ act as penalties on $J_R$ for each constraint, $\bm{\tau} = \{\tau_i\}_{i=1,\dots,n}$ are the constraint thresholds, and $V^{\pi}_{\mathcal{C}} = \{V_{C_i}^\pi\}_{i=1,\dots,n}$ are the expected cost returns.% (defined as in Equation \ref{eq:value_functions}, but over cost functions). % Lagrangian algorithms thus take gradient ascent steps in $\pi$ and descent steps in $\bm{\lambda}$ to trade off safety and task performance. These methods focus on satisfying the constraints using penalties $\bm{\lambda}$ that grows unbounded when constraints are violated. When constraints are satisfied, $\bm{\lambda}$ scale down (to zero), allowing the algorithm to maximize the task objective.

%However, leveraging this formalization for power grids requires formalizing operational constraints (see Section \ref{sec:cmdp_rl2grid}).

\section{RL2Grid Benchmark}
\label{sec:formalization}

RL2Grid considers the general setting of operating a power grid via topology optimization, as well as redispatch and curtailment actions (wrapped within a traditional Gymnasium interface), in order to keep the grid operational over a long horizon---a month of operations divided into 5 minute steps:

\textit{(i) Topology optimization} involves identifying substations where a bus-split action can mitigate the overload by adjusting the grid topology (i.e., how elements are interconnected in the grid). This approach is cost-effective for grid operators as it typically involves simple switch activation.\footnote{There is some uncertainty (and debate) regarding how frequently each component can be switched safely in practice, without degrading the underlying equipment.} However, determining the ``optimal'' topology from the exponential number of possible configurations is typically infeasible using existing optimization-based solvers.

\textit{(ii) Redispatch or curtailment} deals with adjusting the power flow by redispatching or curtailing the power output of fossil and renewable power generators (respectively). However, this method is often economically demanding, as it disrupts the normal operations of third parties controlling the generators and can lead to additional power costs.
% In the following, we discuss the main features of RL2Grid's environments, the TSOs-informed approach to design the action and state spaces and reward function to create a standardized set of tasks. We further present a heuristic module that enables common operations to be incorporated in the transition to a new state, reducing the length of the problem horizon, and introducing safety-focused variants of the tasks by modeling operational constraints.

\begin{wrapfigure}{r}{0.45\textwidth}
    \centering
    \vspace{-5pt}\includegraphics[width=0.37\textwidth]{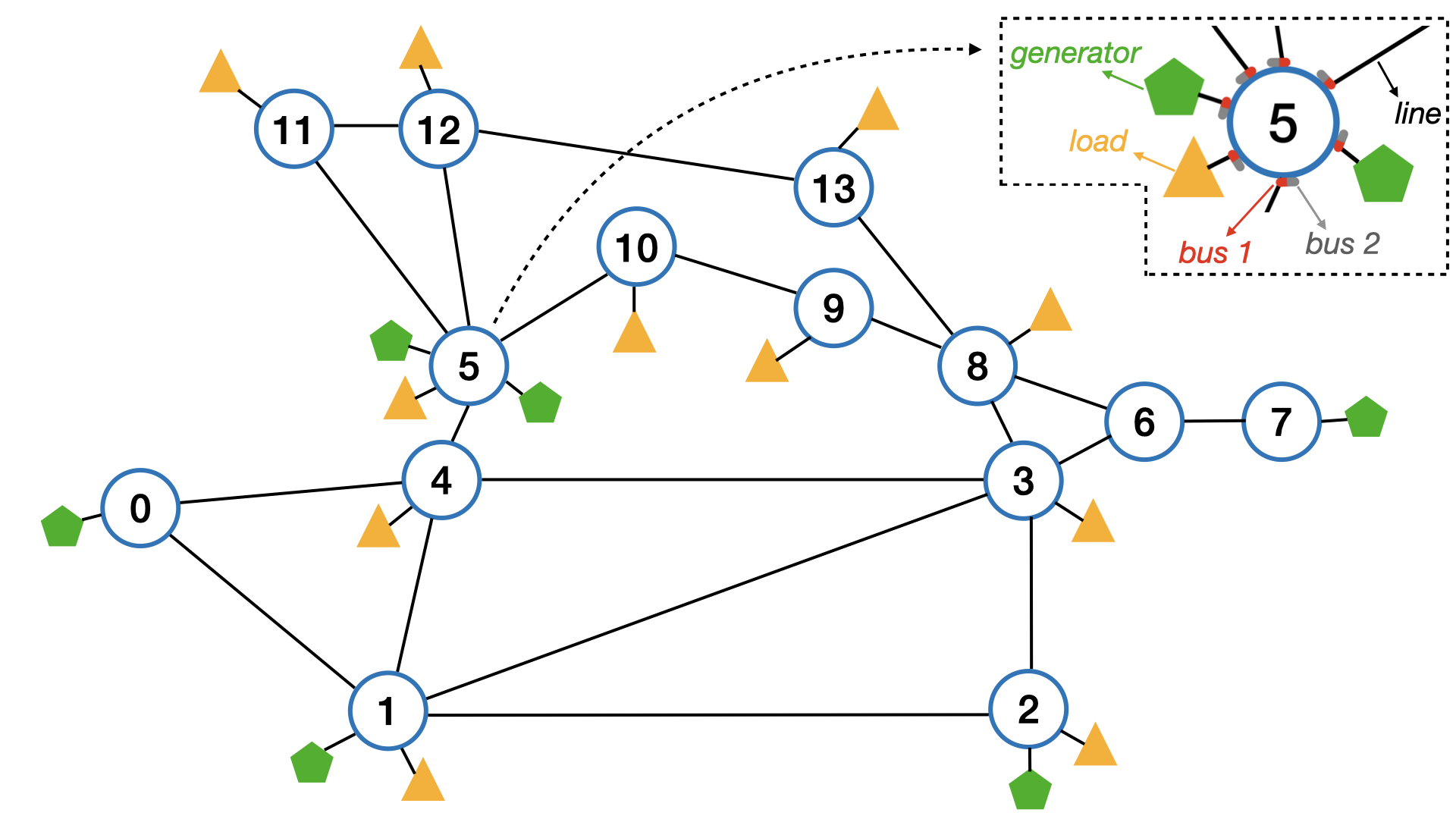} 
    \caption{IEEE 14-bus sample grid.}
    \label{fig:bus14_example}
    \vspace{-5pt}
\end{wrapfigure}
RL2Grid tasks are designed on top of 7 main ``base grids'' from Grid2Op. Each of these grids has a double bus system---every electrical component (i.e., generator, load, and transmission line) has two possible connections within a substation. 
Table \ref{tab:environments} summarizes these base grids, along with the features and the number of components they include. 
These grids present two possible types of contingencies: (i) \textit{Maintenance} (M): Scheduled maintenance events observed by the agent where a line is disconnected and cannot be reconnected until a fixed number of steps (a \textit{cooldown}) has passed; and (ii) \textit{Opponent} (O): Unforeseen events (e.g., weather conditions) that introduce stochasticity by causing a random line to disconnect, entering its cooldown state. The agent does not know about these events in advance and must address the contingencies that such a disconnection might cause in real time. Environments may also include storage units (\textit{Batteries} (B)) that can act as both generators (discharge) and loads (charge). 
\begin{table}[b]
    \centering
    \vspace{-2pt}
    \caption{List of base grid environments and contingencies currently supported by RL2Grid.}
    \label{tab:environments}
    \setlength{\tabcolsep}{3pt}
    \begin{tabular}{lccccccc}
        \toprule
        \textit{\textbf{ID}} & \multicolumn{1}{l}{\textit{\textbf{Maintenance}}} & \multicolumn{1}{l}{\textit{\textbf{Opponent}}} & \multicolumn{1}{l}{\textit{\textbf{Battery}}} & \multicolumn{1}{l}{\textit{\textbf{\# Subs.}}} & \multicolumn{1}{l}{\textit{\textbf{\# Lines}}} & \multicolumn{1}{l}{\textit{\textbf{\# Gens.}}} & \multicolumn{1}{l}{\textit{\textbf{\# Loads}}} \\ \midrule
        \textbf{bus14}                   & $\checkmark$                                      & $\times$                                       & $\times$                                      & 14                                                   & 20                                             & 6                                                   & 11                                             \\
        \textbf{bus36-M}                 & $\checkmark$                                      & $\times$                                       & $\times$                                      & 36                                                   & 59                                             & 22                                                  & 37                                             \\
        \textbf{bus36-MO-v0}             & $\checkmark$                                      & $\checkmark$                                   & $\times$                                      & 36                                                   & 59                                             & 22                                                  & 37                                             \\
        \textbf{bus36-MO-v1}             & $\checkmark$                                      & $\checkmark$                                   & $\times$                                      & 36                                                   & 59                                             & 22                                                  & 37                                             \\
        \textbf{bus118-M}                & $\checkmark$                                      & $\times$                                       & $\times$                                      & 118                                                  & 186                                            & 62                                                  & 99                                             \\
        \textbf{bus118-MOB-v0}           & $\checkmark$                                      & $\checkmark$                                   & $\checkmark$                                  & 118                                                  & 186                                            & 62                                                  & 91                                             \\
        \textbf{bus118-MOB-v1}           & $\checkmark$                                      & $\checkmark$                                   & $\checkmark$                                  & 118                                                  & 186                                            & 62                                                  & 99                                             \\ \bottomrule
    \end{tabular}
\vspace{-10pt}
\end{table}

\textbf{Transition dynamics.} Each scenario in RL2Grid is defined over time series of synthetic but operationally realistic load demand and generation profiles created with ChroniX2Grid~\citep{chronix2grid}.\footnote{We use the time series integrated in Grid2Op's base grids, which consists of months to years' worth of data simulating generation and demand profiles happening over extended periods of time. Due to confidentiality and privacy constraints, up-to-date real-world grid time series data is typically unavailable to use.} At the beginning of each episode, a random time index is sampled from the full-time series to initialize the environment, ensuring that agents are trained under a wide variety of grid conditions and do not overfit to specific temporal patterns. 

From there, at each step, the environment transitions from state $s_t$ to $s_{t+1}$ through a multi-stage process that reflects realistic grid operations. First, stochastic (unforeseen) opponent events (e.g., line faults caused by weather events) are applied, following distributions specific to each base grid. Agent actions (e.g., topology or redispatch changes) are then executed, and invalid actions (e.g., those violating cooldowns) do not have any effect. The environment updates cooldown timers and applies any scheduled maintenance events. The underlying Grid2Op power simulation framework then simulates physical power flows using an AC power flow solver; if the system is infeasible due to islanding or demand shortfall, the episode ends. Otherwise, line overloads are identified, and persistent overloads lasting more than three steps trigger automated disconnections. More details on this are discussed in \citet{grid2op}. The next state is constructed by updating grid variables (e.g., topology, power flows, forecasts), capturing the nonlinear, nonconvex, and stochastic dynamics of real-world power systems.

\textbf{State space.} Agents have access to the state of the power grid at each time step. The state includes common grid features such as production at each generator, load demands, status, capacity, and cooldown of transmission lines, and the current step. Additional features are provided based on the environment's characteristics (e.g., maintenance, opponent events, and/or batteries---see Table~\ref{tab:environments}) and the action space. In the topological case, the state includes the topological vector (an integer vector indicating where each device is connected), the connection status of lines, overflow status, and substation cooldowns. In the continuous case, the state consists of target and actual dispatches, curtailment, and generator ramping limits. These are the main features used by the AC power flow solver to transition the grid to the next state. Due to space limitations, an exhaustive list and description of the features that comprise the state is discussed in \Cref{suppl:state_space}.

\textbf{Action spaces.} Each grid has two types of tasks, depending on the nature of their action space.

(i) \textit{Topology space}: Agents take discrete actions that modify the topology of the substations---disconnecting or reconnecting a line, or changing the bus to which a component is connected.%These actions are virtually free (from an economic cost perspective) since they only involve remotely activating a switch. However, they represent a significant challenge for grid operations, as the number of discrete actions scales exponentially with the number of elements connected to the substation.
\footnote{Human operators currently modify the grid topology manually based on historical behaviors; there is no tractable approach to obtain optimal topology optimization solutions (at scale) as of yet.}
Line switching introduces one discrete action per line, whereas bus reassignments (or ``bus-splitting'') yield an exponentially large number of valid actions depending on the number of elements connected to the substations.
Specifically, the topological action space for a double bus substation composed of $N_{\text{lines}}$ lines, $N_{g}$ generators, $N_{l}$ loads, has size $N = 2 ^{N_{\text{lines}}+N_{g}+N_{l}-1} - 1$ \citep{powrl}.
For instance, substation \#5 in \Cref{fig:bus14_example} has 7 elements, resulting in 63 possible actions, while in the larger \textit{bus36} and \textit{bus118} grids, a single substation can have over 65,000 possible configurations. 

\begin{wrapfigure}{r}{0.5\textwidth}
    \centering
    \vspace{-5pt}\includegraphics[width=0.4\textwidth]{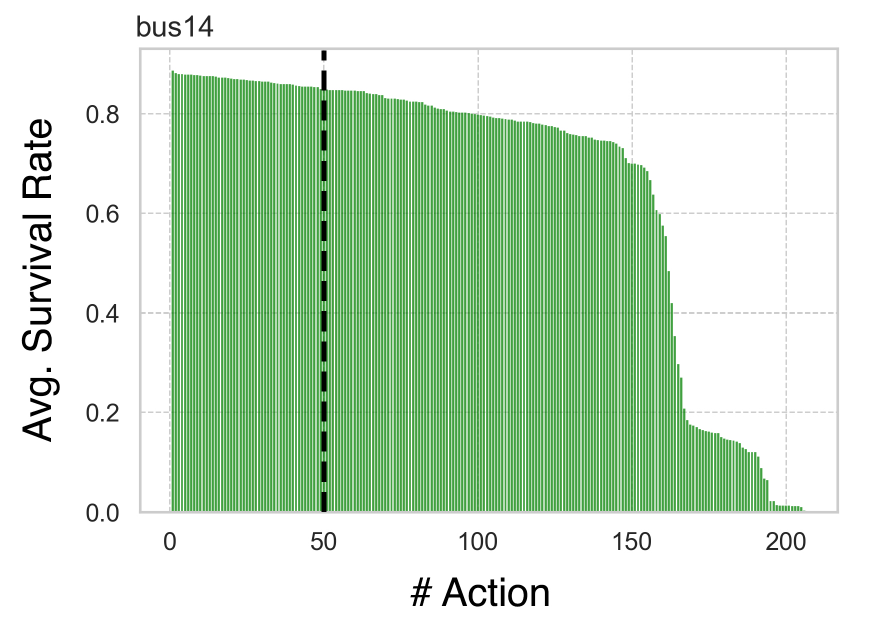} 
    \vspace{-5pt}
    \caption{Action ranking for the \textit{bus14} task. The dashed line separates the difficulty levels (i.e., with 50 and 209 actions).}
    \label{fig:bus14_ranking}
    \vspace{-10pt}
\end{wrapfigure}
Considering the size of the space, we propose ``difficulty levels'' in which the action space has an increasing number of topology actions. We selected these action spaces through extensive simulations (72 hours on the computer cluster detailed in \Cref{sec:experiments}) by ranking the full action space based on the \textit{survival rate} for the grid. 
This rate represents the number of steps that each action maintains the grid in normal conditions over an episode, and is defined as \textit{the normalized number of steps for which an action does not cause a grid collapse} (because the total demand is not satisfied or parts of the grid become disconnected). In detail, we uniformly sampled topological actions, and after ordering them by highest survival rate, we took the first $N_{\text{actions}}$ from the ordered space, where $N_{\text{actions}}$ increases with each difficulty level. For example, \Cref{fig:bus14_ranking} shows the ranking for the \textit{bus14} grid, highlighting how suitable all the topological actions are to address grid contingencies (and how easier levels contain actions that are more likely to address a contingency). To further motivate our method, we visually analyze the impact of the resultant action spaces in \Cref{suppl:env}, where we also summarize the difficulty levels and the size of their discrete action spaces. Considering these levels, RL2Grid has a total of 32 topology-based tasks.

(ii) \textit{Redispatching and curtailment space}: Agents take continuous actions changing how power generation is scheduled. Unlike topology actions, which only involve remotely activating a switch, these actions are not free. Economic costs arise from altering the planned generation schedule of power plants, increased fuel costs, and financial compensation for renewable energy producers. 
Redispatching actions apply to fossil fuel-based generators, while curtailment actions apply to renewable energy-based generators. 
Batteries, if present, are also considered generators and come with continuous actions for charging/discharging operations. % (e.g., \textit{bus118-MOB-v0} has 62 generators and 7 batteries since its continuous action space has size 69).
This action space is relatively tractable for RL algorithms since it involves one continuous action per generator (i.e., $N = N_g$). 
Thus, we present a total of 7 continuous action-based training environments (one per base grid).

\textbf{Reward.} The reward design is informed by TSOs' mandate to satisfy real-world grid operation requirements:~promote long-term safety and efficiency by rewarding grid survival and penalizing unsafe or costly actions (in terms of economic costs). At each step $t$, the reward an agent gets is $R_t = \alpha R_{\text{survive},t} + \beta R_{\text{overload},t} + \eta R_{\text{cost},t}$, where the weights are evaluated in \Cref{suppl:hyperparameters}. Specifically, the agent gets a cumulative positive constant $R_{\text{survive},t}$ for each step, normalized by the total length of a training episode ($\in [0, 1]$). The overload and cost rewards are defined as: 

\textit{(i) Overload}: Penalizes line overloads and disconnections, and rewards available line capacity based on the difference between line flows and capacity limits. In unconstrained settings, disconnected lines incur a fixed penalty. This is more formally defined as:
    \[
    R_{\text{overload},t} = \sum_{\ell \in \mathcal{L}} \left[ \max\left(0, \frac{P_{F,\ell,t} - P_{F,\ell}^{\max}}{P_{F,\ell}^{\max} + \epsilon} \right) - \mathbbm{1}(\ell \text{ is disconnected}) \right],
    \]
where $P_{F,\ell,t}$ is the power flow on line $\ell$ at time $t$, $P_{F,\ell}^{\max}$ is its capacity limit, $\epsilon$ is a small constant to avoid divisions by 0, and the indicator function returns 1 if the line is disconnected. This term is then normalized to lie within $[-1, 1]$. 

\textit{(ii) Cost}: Penalizes redispatching or curtailment actions based on deviations from planned dispatch schedules and energy losses. This is defined as:
\[R_{\text{cost},t} = - \left[ \left(P_{G,t} - P_{D,t}\right) + \vert c_{\text{redisp}, t}\vert + \vert P_{\text{storage}, t}\vert \right] c_{\text{marginal}, t},\]
where $P_{G,t}$ and $P_{D,t}$ denote the total power generated and total demand consumed at time $t$, respectively, with their difference representing transmission losses, $c_{\text{redisp}, t}$ corresponds to the redispatched power (i.e., the absolute deviation from scheduled generator setpoints), and $P_{\text{storage}, t}$ represents the power exchanged with storage units. All cost components are scaled by the marginal generation cost $c_{\text{marginal}, t}$, defined as the cost per MWh of the most expensive generator currently producing power. This value is also normalized to lie in the range \([-1, 0]\).

\subsection{Heuristic-guided Transitions}
\label{sec:heuristic}

\begin{figure}[b]
    \centering
    \includegraphics[width=1.0\textwidth]{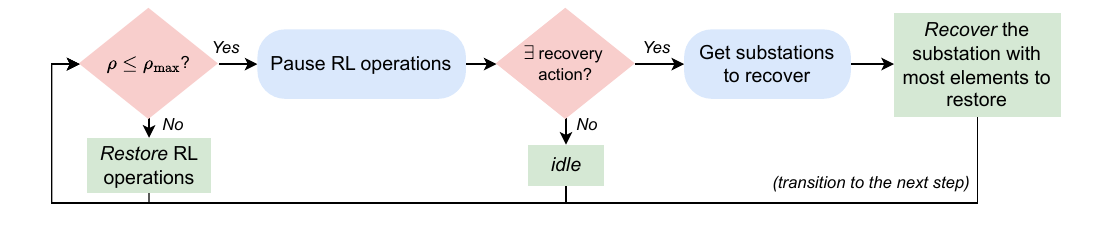}
    \vspace{-10pt}
    \caption{Recovery heuristic combining L2RPN strategies and operator expertise. The agent acts under risk (e.g., line overloads); otherwise, the heuristic incrementally restores the original topology.}
    \label{fig:heuristic}
    \vspace{-5pt}
\end{figure}

To reduce the problem horizon and (potentially) improve learning stability and sample efficiency, RL2Grid incorporates two expert-informed heuristics that modify the transition dynamics described above. These heuristics emulate human operator behavior: they suppress agent actions when the grid is stable and allow the agent to try to recover normal operations when contingencies occur.

The \textit{idle heuristic (I)} is triggered when all line loadings are below a safety threshold set to $95\%$ of their capacity. In this case, the agent’s control is suspended and replaced by no-op actions. The \textit{recovery heuristic (R)}, shown in \Cref{fig:heuristic}, activates when the grid is safe but the topology differs from its original configuration. It incrementally restores the original topology, modifying at most one substation per step, and defaults to idle once recovery is complete.

These heuristics modify the environment so that each agent action initiates a sequence of $n$-step heuristic-guided transitions during which rewards accumulate. While our heuristics encode operational priors from winning L2RPN competitors \citep{marot2020learning, marot2021learning}, our methods have the benefit of being embedded directly in the environment for compatibility with standard RL libraries and are systematically benchmarked. We believe this design balances expert-in-the-loop guidance with the learning process, enabling more sample-efficient training and improved policy performance in realistic, safety-critical scenarios.

\subsection{Fostering Safe Operations via Constrained RL}
\label{sec:cmdp_rl2grid}

While prior works such as L2RPN have not incorporated constrained formulations to foster safety~\citep{marot2020learning, marot2021learning, marot2022perspectives}, RL2Grid introduces CMDP-based tasks that reflect two key classes of safety violations faced by system operators as constraints.

In the \textit{load shedding and islanding (LSI)} case, unsatisfied demand and islanding trigger a positive cost. Let us denote the total demand and generation at time $t$ as $P_{D,t}$ and $P_{G,t}$, respectively, and define $L_t = \mathbbm{1}(P_{G,t} < P_{D,t}), \quad I_t = \mathbbm{1}(N_{I,t} > 0),$
where $N_{I,t}$ is the number of disconnected components. The LSI cost is $C_{\text{LSI}}(t) = L_t + I_t$,
with a zero cumulative threshold $\sum_t C_{\text{LSI}}(t) = 0$ to model a hard safety constraint. For \textit{transmission line overload (TLO)}, a positive cost occurs upon thermal overloads and line disconnections. Let us denote the power flow on line $\ell$ as $P_{F,\ell,t}$, its capacity as $P_{F,\ell}^{\max}$, and define $O_{\ell,t} = \mathbbm{1}(P_{F,\ell,t} > P_{F,\ell}^{\max}), \quad D_{\ell,t} = \mathbbm{1}(\ell \text{ is disconnected})$, where $D_{\ell,t}$ excludes scheduled maintenance or opponent-driven disconnections. The TLO cost is $C_{\text{TLO}}(t) = \sum_{\ell \in \mathcal{L}} (O_{\ell,t} + D_{\ell,t})$, with a cumulative threshold $\sum_{t=0}^T C_{\text{TLO}}(t) < \tau$ to model this as a ``soft'' constraint.

These constraints are applied across all 32 topology-based environments, resulting in 64 additional constrained variants.\footnote{Constraints are compatible with redispatching tasks but are primarily evaluated on topology due to their operational relevance and the complexity of associated actions.} By modeling these critical safety violations, RL2Grid enables the development and benchmarking of safe RL methods for real-world grid operations.

\section{Experiments}
\label{sec:experiments}

We evaluate the performance of baseline RL algorithms that typically serve as building blocks for more complex algorithms in representative RL2Grid tasks.
% \begin{itemize}[noitemsep]
%     \item \textit{DQN} \citep{dqn} approximates the $Q$-function, using an $\epsilon$-greedy policy at training time. Due to its value-based nature, a DQN agent can only consider discrete (topological) actions.
%     \item \textit{PPO} \citep{ppo} and \textit{Lagrangian PPO} (LagrPPO) \citep{LagrangianPPO} directly approximates a policy by learning its parameters using a computationally tractable clipped objective. By learning different probability distributions, a PPO agent can deal with continuous (redispatching) and discrete (topological) actions. The lagrangian version is used in the constrained topological environments as it represents a popular and good performing baseline for CMDP-based tasks \citep{ijcai2021constraints}.
%     \item \textit{SAC} \citep{sac} uses different networks to learn a policy and two value functions that mitigate positive bias in value estimates. An SAC agent can deal with the same actions as PPO.
%     \item \textit{TD3} \citep{td3} is similar to SAC, but uses multiple networks to learn a deterministic policy and can commonly deal with continuous actions.
% \end{itemize}
% ---that is DQN \citep{dqn}, PPO \citep{ppo}, \citep{LagrangianPPO}, SAC \citep{sac}, TD3 \citep{td3})---
In particular, we test: (i) (double) DQN \citep{doubledqn}, PPO \citep{ppo}, and SAC \citep{sac} and their heuristic versions on the discrete topological action space for the \textit{bus14, bus36-MO-v0, bus118-M, bus118-MOB-v0} tasks over most levels of difficulty; (ii) PPO, SAC, and TD3 \citep{td3} in the continuous redispatching action space of these environments; (iii) and the Lagrangian version of PPO, LagrPPO \citep{LagrangianPPO}, in the two constrained versions (LSI and TLO) of the topological \textit{bus14} task. We consider these tasks to be representative, as they provide sufficient empirical evidence of the current performance of the baselines in power grid operations. % Our experiments address the following key questions: (i) \textit{Can commonly-used model-free RL methods deal with high-dimensional power network operations?} (ii) \textit{What is the impact of integrating existing task-level knowledge as a heuristic-guided policy within these real-world tasks?} (iii) \textit{How difficult is it to consider constraints when training an RL agent for power grid operations?}

\textbf{Implementation and Data Collection.}
Data collection is performed on Xeon E5-2650 CPU nodes with 256GB of RAM, using CleanRL-based implementations for the baselines \citep{huang2022cleanrl} and the hyperparameters---selected via grid search---in \Cref{suppl:hyperparameters}. If not specified otherwise, the results show the average survival smoothed over 500 episodes of 10 runs per method, with shaded regions representing the $95\%$ confidence intervals. As described above, the survival denotes the normalized number of time steps the grid remains operational over an episode, with a survival rate of 1 indicating one month of successful grid operations. We set a strict time limit on the nodes used for data collection, set to 48 hours for each individual run. % As such, some setups run for more steps than others. For example, the heuristic-based experiments are more computationally demanding since they often check and compute extra recovery actions. 
The experiments in this work (excluding the hyperparameter search) required a total of $>$180,000 CPU hours to execute, and
% for our experiments, 
\Cref{suppl:env_impact} addresses the associated environmental impact and our efforts to offset estimated CO$_2$ emissions.

\begin{figure}[b]
    \centering
    \vspace{-5pt}\includegraphics[width=1.0\textwidth]{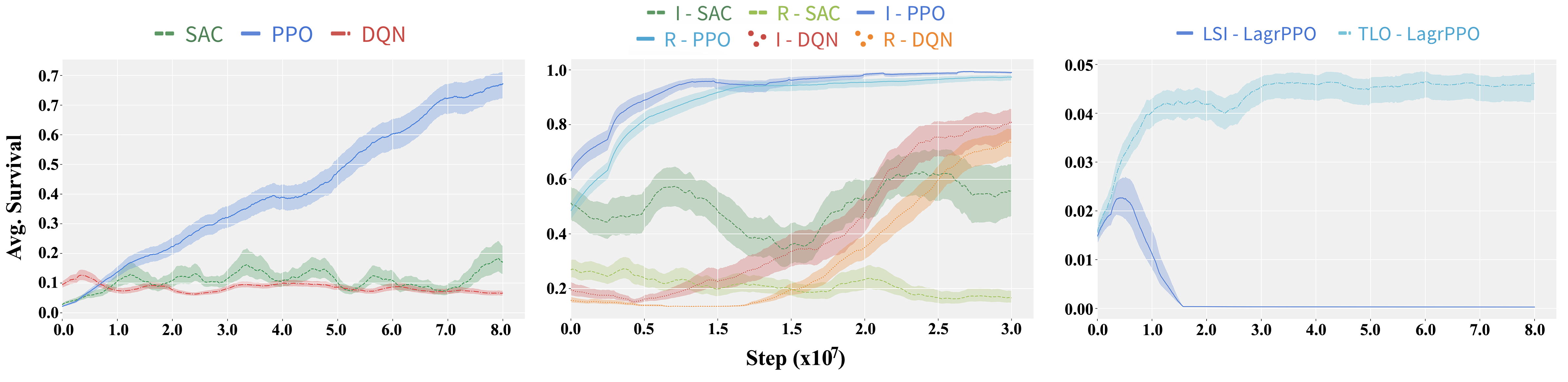}
    \vspace{-8pt}
    \caption{Average survival of the baselines (left), heuristic versions (center), and constrained variants (right) on the bus14 task with topological actions at difficulty 0 (higher values are better). 
    }
    \label{fig:bus14_paper}
    \vspace{-5pt}
\end{figure}

\begin{table*}[t]
\centering
\label{tab:new_results}
\caption{Average survival rate (higher is better) in a subset of difficulty 0 tasks with topological actions for the baselines: idle (\textit{I}), MILP optimization (\textit{MILP}); and RL algorithms: vanilla RL (\textit{V}), and the heuristic versions (\textit{I}, \textit{R}) for DQN, PPO, and SAC.}
\setlength{\tabcolsep}{5pt}
\begin{tabular}{llccccccccccc}
\toprule
                       &                & \multicolumn{2}{c}{\textbf{Baseline}} & \multicolumn{3}{c}{\textbf{DQN}}                        & \multicolumn{3}{c}{\textbf{PPO}}                        & \multicolumn{3}{c}{\textbf{SAC}}                        \\ \midrule
\textbf{Env.}          & \textbf{Diff.} & \textit{I} & \textit{MILP} & \textit{V} & \textit{I} & \textit{R} & \textit{V} & \textit{I} & \textit{R} & \textit{V} & \textit{I} & \textit{R} \\ \midrule
\textbf{bus14}         & \textit{0}     & 0.18          & 0.06              & 0.07       & 0.86       & 0.74       & 0.74       & \textbf{0.99}       & 0.97       & 0.17       & 0.56       & 0.16       \\
\textbf{bus36-MO-v0}   & \textit{0}     & 0.03          & 0.06              & 0.04       & 0.14       & 0.19       & 0.06       & 0.17       & \textbf{0.29}       & 0.01       & 0.10       & 0.13       \\
\textbf{bus118-MOB-v0} & \textit{0}     & 0.10          & 0.05              & 0.07       & 0.19       & 0.27       & 0.04       & 0.18       & \textbf{0.28}       & 0.01       & 0.15       & 0.19       \\ \bottomrule
\end{tabular}
\end{table*}

% \begin{table*}[t]
% \centering
% \label{tab:new_results}
% \vspace{-5pt}
% \caption{Average survival rate (higher is better) for vanilla baselines (\textit{V}) and the heuristic versions (\textit{I}, \textit{R}) in a subset of difficulty 0 tasks with topological actions for DQN, PPO, and SAC.}
% \setlength{\tabcolsep}{5pt}
% \begin{tabular}{lllllllllll}
%  \toprule
%                        &                & \multicolumn{3}{c}{\textbf{DQN}}                        & \multicolumn{3}{c}{\textbf{PPO}}                        & \multicolumn{3}{c}{\textbf{SAC}}                        \\ \midrule
% \textbf{Env.}          & \textbf{Diff.} & \textit{V} & \textit{I} & \textit{R} & \textit{V} & \textit{I} & \textit{R} & \textit{V} & \textit{I} & \textit{R} \\ \midrule
% \textbf{bus36-MO-v0}  & \textit{0}     & 0.04          & 0.14               & 0.19               & 0.06          & 0.17               & 0.29               & 0.01          & 0.10               & 0.13               \\
% \textbf{bus118-M}      & \textit{0}     & 0.06          & 0.17               & 0.18               & 0.07          & 0.13               & 0.18               & 0.15          & 0.18               & 0.19               \\
% \textbf{bus118-MOB-v0} & \textit{0}     & 0.07          & 0.19               & 0.27               & 0.04          & 0.18               & 0.28               & 0.01          & 0.15               & 0.19               \\ \bottomrule
% \end{tabular}
% \end{table*}%
\begin{table}[t]
\centering
\label{tab:preliminary_results}
\caption{Average survival rate (higher is better) of the grid in a subset of tasks with continuous redispatching actions obtained by an idle baseline (\textit{I}) and RL-based vanilla baselines (\textit{V}) for PPO, SAC, and TD3.}
\setlength{\tabcolsep}{5pt}
\begin{tabular}{llcccc}
\toprule
\textbf{Env.}          & \textbf{Diff.} & \textit{I} & \textbf{PPO} (\textit{V}) & \textbf{SAC} (\textit{V}) & \textbf{TD3} (\textit{V}) \\ \midrule
\textbf{bus14}         & \textit{0}     & 0.00       & \textbf{0.17}                    & 0.01                    & 0.06                    \\
\textbf{bus36-MO-v0}   & \textit{0}     & \textbf{0.08}       & \textbf{0.08}                    & 0.02                    & 0.01                    \\
\textbf{bus118-MOB-v0} & \textit{0}     & 0.11       & \textbf{0.25}                    & 0.08                    & 0.07                    \\ \bottomrule
\end{tabular}
\vspace{-5pt}
\end{table}

\textbf{Results.} We indicate with \textit{V} the ``vanilla'' baselines, and with \textit{I} and \textit {R} the experiments with the heuristics described in \Cref{sec:heuristic}. %For these topological cases, we report the results for the first level of difficulty (i.e., level 0, considering 50 discrete actions), and Table 3 shows the results for the redispatching action spaces.%\footnote{We recall the redispatching case only has one level of difficulty since it only considers one action for each generator.} 
Overall, the RL baselines struggle to deal with the real-world complexities of power grid operations. As expected, we also notice that the heuristic-guided transitions reduce the problem complexity and typically achieve higher performance, despite being not nearly sufficient to operate complex grid setups for long periods of time. 

\Cref{fig:bus14_paper} compares the training performance of the baseline algorithms, the heuristic versions, and the constrained variants on the topological \textit{bus14} grid at difficulty level 0. Among the unconstrained baselines, only \emph{PPO successfully learns an effective policy} in this relatively simple environment. However, incorporating human-informed heuristic operations leads to notable improvements in both performance and sample efficiency across all methods. With heuristic augmentation, PPO achieves good long-term control, while DQN and SAC also exhibit strong performance.
Interestingly, our results indicate that \emph{performing idle operations in smaller grid domains---rather than reverting to the original topology---can improve RL performance}. In contrast, introducing constraints significantly increases task difficulty: agents are penalized for violations, and LagrPPO struggles even in this basic setting, failing to learn effective control and frequently exceeding constraint thresholds (see \Cref{suppl:results} for details).
Finally, Tables 2 and 3 report the average survival at convergence for the unconstrained baselines in both topological and redispatching tasks.\footnote{Due to space constraints, we report only average performance. Full results and training curves, including the \textit{bus118-M} task and higher difficulty levels, are available in \Cref{suppl:results}.} For topology control, we evaluate two traditional baselines: an ``idle'' policy (I) and the MILP-based agent from Grid2Op \citep{grid2op_milp}, which minimizes line overloads via topological actions under DC power flow approximations.\footnote{This approximation is necessary due to the intractability of AC formulations \citep{marot2020learning, marot2021learning}.} For redispatching, only the idle baseline is considered, as no built-in agent is provided in Grid2Op. Notably, even model-free RL-based agents consistently outperform these traditional methods across both settings.

\textit{These performance results motivate the need for further advancements in RL algorithms that can contend with the complex dynamics and aleatoric uncertainty, long-horizon goals, and hard physical constraints of real-world tasks such as power grid operations}. %By providing common ground for the community, we hope to foster further research on these fronts.

\begin{figure}[t]
    \centering
    \includegraphics[width=1.0\textwidth]{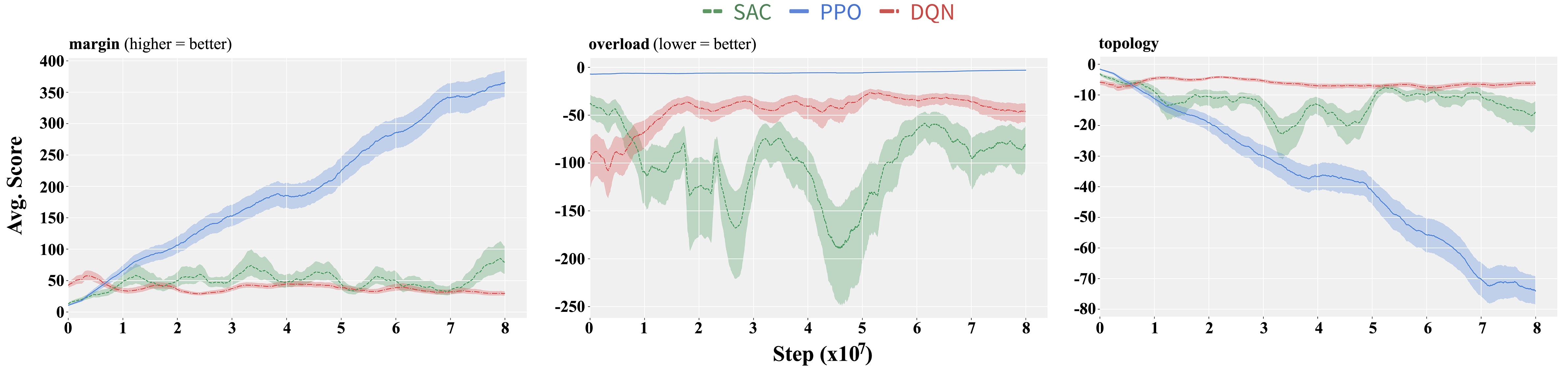}
    \vspace{-10pt}
    \caption{Average score for key operational components across vanilla baselines. The first column represents line margins, where higher values indicate better contingency management. The second column tracks overload penalties, with lower values reflecting improved grid stability. The third column captures topology modifications, showing the extent to which agents reconfigure the grid.}
    \label{fig:bus14_paper_analysis_v2}
    \vspace{-5pt}
\end{figure}

\textbf{Performance analysis.} \Cref{fig:bus14_paper_analysis_v2} analyzes how the baseline algorithms are learning to control the grid.\footnote{A similar analysis for the heuristic and constrained experiments is available in \Cref{suppl:performance analysis}.} In particular, we quantify three distinct operational metrics---maring, overloads, and topology---each defined and shown (with $95\%$ confidence intervals) as a scalar-valued ``score" signal derived from the agent’s interaction with the environment:
\begin{itemize}[noitemsep]
    \item \textit{Margin} (first column): Represents the cumulative available margin across all transmission lines, where disconnections are penalized and lower power usage is rewarded. It is defined per-step $t$ as: $\sum_{\ell \in \mathcal{L}} m_{\ell, t}$, where $m_{\ell, t} = \max\left(0, P_{F, \ell}^{\max} - |P_{F, \ell, t}| \right)$, if line $\ell$ is connected, $m_{\ell, t} = -1$ if not.
    Higher values indicate that the agent maintains greater flexibility to handle contingencies. Overall, we find that successful agents tend to maximize line margins.
    
    \item \textit{Overloads} (second column): Shows the overload component $R_{\text{overload,t}}$ as defined in \Cref{sec:formalization}. Lower values imply that the agent maintains power flows within safe operational limits, which is a key characteristic of effective policies. Unsurprisingly, the unconstrained agents violate the overload constraints. Perhaps surprisingly, the constrained versions actually exhibit even more severe violations of these constraints, which we hypothesize is due to their overall difficulty in optimizing performance in the constrained environment.

    \item \textit{Topology} (third column): Quantifies deviations from the initial grid configuration (where all elements are connected to the first bus). It is defined per-step $t$ as: $-d(G_t, G_0)$, where $G_t$ denotes the grid topology at time $t$, and $d(G_t, G_0)$ is the Hamming distance of the topology at time $t$ from the initial topology $G_0$. This metric quantifies the extent of grid reconfiguration. Negative values indicate more significant structural interventions, which often correlate with better operational performance in learned policies.
    A value close to $0$ suggests minimal changes, while higher values indicate significant topological modifications. We find that successful agents tend to actively reconfigure the grid to optimize operations.
\end{itemize}

\section{Related Work} 

Recent attempts to develop sequential decision-making in power system operations benchmarks often focus on small-scale problems and/or simplified setups \citep{rl_powersystem}. Examples include \citet{pygrid} for microgrids, CityLearn for demand response and urban energy management \citep{citylearn}, and gym-ANM for small electricity distribution networks \citep{gym_anm}. 
RL environments for electric vehicle (EV) charging and electricity markets have also been introduced \citep{rl_powersystem2}. Recently, SustainGym spanned diverse tasks ranging from EV charging to carbon-aware data center job scheduling \citep{sustaingym}.
The ARPA-E GO Competition provides a large-scale benchmark for power grid operations, but is more-so geared towards offline optimization approaches \citep{go2023homepage}.  On the methodological side, recent contributions in the field include works on cascading failure mitigation, demand response optimization, and real-time grid control using RL \citep{rev3, rev1, rev2}. Nonetheless, these works are more geared towards methodological advancements rather than proposing a benchmark. For this reason, we refer the reader to recent reviews for details on RL applications in power grid operations \citep{rlgrid_survey, su2024review}.

\paragraph{Relationship with Grid2Op.} Grid2Op is an open-source power simulation framework designed by RTE France to model power grid controllers \citep{grid2op}. %It includes realistic non-linear dynamics, uncertainty deriving from time-varying renewable energy sources, and the massive amount (combinatorially large number) of grid configurations that exist even in moderately-sized grids.
It simulates grid operations, requiring that grids work for long horizons in a way that is robust to contingency events, %(e.g., disconnections, maintenance, and stochastic or ``adversarial'' overloads caused by extreme weather conditions), 
as well as adhering to physical and operational constraints. For the latter, Grid2Op models: (i) cooldown periods to prevent immediate reconnection of disconnected lines, and limits on the frequency of actions on the same line to avoid asset degradation; (ii) limited thermal capacity of transmission lines; (iii) ramp rates on generators that restrict how much power generation can change between time periods; and (iv) adherence to AC power flow constraints. %Moreover, using packages like chronix2grid \citep{chronix2grid}, Grid2Op provides realistic time series data for load demands and generator outputs under ideal conditions. 
Grid2Op has been primarily used as the base for the L2RPN competitions \citep{marot2020learning,marot2021learning,marot2022learning}. While RL methods have been used for L2RPN, they fail to provide a common ground to foster advancements in the field---each method uses custom input features and action spaces of (very) limited size, often without providing sufficient evidence on how and why these spaces were considered. 
Hence, to date, there is no standardized solution that allows RL researchers to easily get started in this field and compare over an established benchmark.\footnote{\Cref{suppl:l2rpn_comparison} further discusses the relationship between L2RPN and RL2Grid.}
\section{Tackling the Challenges of Power Grids with RL}
\label{sec:discussion}

Applying RL in power grids presents numerous open problems, each offering significant opportunities for advancing both grid operations and RL methodologies \citep{marot2022perspectives}. While we address a subset of these challenges via our work, there remains ample room for future work.

\subsection{Relevant RL Methodologies} RL has the potential to be beneficial in addressing open grid problems. There are also potential risks (e.g., with respect to safety, reliability, and robustness) that are important to address.
Here, we summarize interesting avenues for future research. 

\textbf{Safe RL.} Safety is paramount in power grid operations. 
Safe RL methods aim to ensure that learning and control policies adhere to strict safety constraints, preventing actions that could lead to blackouts or equipment damage \citep{grid2op}. Ensuring safety while optimizing performance is a critical area of research \citep{garciasafety, epsretrain}. In particular, incorporating novel algorithms based on our CMDP representations can be particularly beneficial for ensuring that solutions adhere to physical and operational limits \citep{constrained_rl, AAMAS2023_violationpenalty}.

\textbf{Human-in-the-loop.} 
Effective grid management requires human expertise and intervention. Incorporating human supervision, interaction, and feedback into RL systems allows for a synergistic approach where human operators and AI work together to optimize grid operations \citep{marot2022perspectives}. This collaboration can enhance decision-making and build trust in AI-driven solutions.

\textbf{Hierarchical control and multi-agent RL.} 
Power grids operate across multiple hierarchical levels, from individual substations to entire regions. Effective coordination within and across these levels is crucial for maintaining efficient and reliable operations. Hierarchical RL methods can be developed to manage multi-level control tasks, in a way that addresses the scale and complexity of grid operations \citep{hierarchical_rl, multiagent2023control}. Another promising direction is the use of multi-agent representations. Given the vast and distributed nature of power grids, scalability can be enhanced by dividing the grid into distinct areas or agents, each responsible for its own operations. Multi-agent RL (MARL) frameworks can enable these agents to learn and coordinate actions \citep{marl_benchmark, marlfactorization2024}, to improve overall grid performance while managing local contingencies more effectively.

\textbf{Robust RL.} The integration of renewable energy sources introduces significant variability and uncertainty into power grids, leading to non-stationary environments. RL algorithms need to adapt to these evolving dynamics to ensure stable and efficient grid operations despite fluctuating supply and demand profiles. 
Handling non-stationarity is thus a critical research direction \citep{robust_rl, vfs2023}.

\textbf{Model-based RL.} Model-based RL methods leverage models of the grid dynamics to improve learning efficiency and policy performance. These methods can provide more accurate predictions and better generalize across different scenarios, leading to faster and more robust solutions \citep{model_based_rl}. Additionally, the AlphaZero algorithm, which combines tree search with deep learning, has shown remarkable success in games like chess and Go and could offer new strategies for handling complex, sequential decision-making tasks with high-dimensional spaces \citep{liu2023real}.

\textbf{Better representations.} Improving model representations for RL in power grids can also lead to more efficient learning and better policy performance. 
Leveraging graph neural networks (GNNs) offers a potential avenue for advancement. Power grids can be naturally represented as graphs, with nodes representing buses and edges representing transmission lines. GNNs can effectively model these structures, capturing the spatial and topological dependencies inherent in power grids. Integrating GNNs with RL algorithms can enhance the representation and learning of grid dynamics.

\textbf{Non-RL approaches.}~While RL holds great promise, it is also essential to consider non-RL approaches such as optimization solvers, which are relevant particularly for problems with well-defined optimization objectives and constraints. In addition, exploring hybrid methods that combine RL with traditional optimization techniques can yield powerful tools for complex grid management tasks.

\subsection{Improving Realism of Power Grid Environments} It is important to acknowledge that RL2Grid is only a first step. Notably, developing ``last-mile'' deployable solutions will require further improvements in the realism of power grid environments, which we now discuss.

\textbf{Scalability.} 
Realistic power systems akin to those managed by RTE France and other transmission system operators
may capture hundreds to thousands of buses. To ensure that RL solutions are applicable to real-world scenarios, improving the size and scale of grid environments is essential.

\textbf{Real data.} Grid2Op (and thus, RL2Grid) relies on realistic but synthetic data, which already provide significant challenges for RL. 
After scaling up RL to deal with the challenges provided by RL2Grid, future environments should (in a way that is cognizant of privacy issues)  publicly release real or more realistic synthetic grid data to design to bridge the gap with real power grid operations.

\textbf{N-1 security.} Grid operators must ensure the system can withstand failure of any single component. 
Rather than modeling failures via random opponents, environments should handle this exhaustively and/or through adversarial agents tailored specifically to the method being tested.

\textbf{Topology vs.~redispatch.} Different grid operators handle the relationship between redispatch and topology optimization differently (e.g., some co-optimize these processes, whereas others prefer to handle them separately). Future benchmarks should reflect this heterogeneity in how different power grids are managed. 
Moreover, Grid2Op's current approach of disconnecting lines after unaddressed overloads does not fully capture real-world practices, where operators attempt to prevent overheating at all costs. Incorporating more realistic consequences for unaddressed overloads, such as system costs, can improve the fidelity of benchmarks. %simulations.
Additionally, grid operators cannot switch every element to every busbar, and there are limits on the number of connected components per substation. Reflecting these constraints can lead to more practical and applicable RL solutions. Storage assets also play an increasingly important role in grid operations. Future benchmarks should accurately model storage and clarify the extent of control grid operators have over these assets.

\textbf{Phase-shift transformers.} Phase-shift transformers, currently modeled as integer variables in the action space, should be represented more accurately to reflect their operational impact. Maintenance activities also vary significantly, with Type A involving physical presence at the site and Type B allowing remote interventions. Differentiating these types of maintenance activities in benchmarks can provide a more accurate representation of real-world constraints.
\section{Conclusions}
\label{sec:conclusions}
Power grids are essential in combating climate change, requiring a transition to low-carbon energy and enhanced resilience against climate-induced extremes. The integration of VRE sources introduces complexities and uncertainties in grid operations, posing significant challenges for human operators and traditional solvers.
Our work aims to foster progress towards
% addresses 
these challenges by introducing RL2Grid, a benchmark
% ing framework 
designed to bridge the gap between current grid management practices and RL research. RL2Grid provides a standardized interface for power grid environments, featuring common rewards, state spaces, action spaces, and safety constraints across a pre-designed set of diverse and complex grid tasks in order to provide a common ground for monitoring and promoting progress. 
% This framework extends the CleanRL codebase and incorporates a heuristic module to improve performance and sample efficiency.
We perform a comprehensive evaluation of the performance of popular baselines on RL2Grid tasks, including versions augmented with domain-informed heuristics aimed at improving performance and sample efficiency, and find that there is still significant room for improvement in the performance of these methods. 
%
%By offering a standardized platform for RL research in the context of power grids, 
RL2Grid aims to accelerate algorithmic innovation 
% and enhance 
% the reliability and efficiency of grid operations 
towards improving power grid operations
amidst the evolving challenges posed by climate change.

%%%%%%%%%%%%%%%%%%%%%%%%%%%%%%%%%%%%%%%%%%%%%%%%%%%%%%%%%%%%

\section*{Acknowledgments}
This work was supported in part by the MIT Climate Nucleus Fast Forward Faculty Fund Grant Program, the AI2050 program at Schmidt Sciences (Grant G-24-66236), and the MIT-IBM Watson AI Lab.

\bibliographystyle{apalike}
\bibliography{biblio}

%%%%%%%%%%%%%%%%%%%%%%%%%%%%%%%%%%%%%%%%%%%%%%%%%%%%%%%%%%%%

\newpage
\appendix

\appendix
\numberwithin{equation}{section}
\numberwithin{figure}{section}
\numberwithin{table}{section}

\section{Relationship of RL2Grid to L2RPN tasks and solutions}
\label{suppl:l2rpn_comparison}

In this section, we clarify the relationship of the tasks presented within RL2Grid, as well as the baseline methods evaluated, to the tasks and solutions presented within the L2RPN competition series. 

We remind that our work builds on Grid2Op to provide a benchmark with standardized tasks, state and action spaces, rewards, and a safe (constrained) formalization, as well as comprehensive evaluation of common baselines inspired by L2RPN. These are critical to provide a common basis for assessing advances in RL methods \citep{marl_benchmark} as well as to improve accessibility to RL practitioners who may have limited prior knowledge of power systems.

\textbf{Tasks.} RL2Grid employs all the main Grid2Op ``base environments'' (which are likewise employed in L2RPN). However, the solutions developed for L2RPN relied on different customized components. Every competition relied on different time series, making effective comparisons far from trivial. For these reasons, on top of the standardization proposed in our work (see Section~\ref{sec:formalization}), we have made some underlying changes to the base environments to better reflect the current and future challenges of RL research. Examples include (i) episodes with longer horizons (i.e., an RL2Grid episode models a month of grid operations, $\sim$8000 steps, compared to weekly episodes of most prior work); (ii) making the tasks as uniform as possible (i.e., by integrating curtailment operations in all Grid2Op tasks); (iii) enabling simulation steps inside the Gymnasium interface (a feature added in our code revision, which is not currently available in Grid2Op). These decisions were driven by our goal of ensuring that our benchmark is accessible, standardized, and provides a clear starting point for researchers who may not be familiar with the nuances of these competitions and power grids.

\textbf{Baselines.} Due to the different choices of input features and action spaces considered by different methods submitted to the L2RPN challenges, it was not possible to directly benchmark these specific methods on the RL2Grid tasks. However, the baselines chosen are representative of the methods submitted to past L2RPN competitions, in addition to representing commonly-used methods within the RL community as a whole.
In particular, within the L2RPN submissions, a common approach was to incorporate heuristics. These heuristics varied significantly between methods and pushed us to design one that mimicked human operations in real grid operations. We developed this heuristic in collaboration with power system operators who have contributed to our work, incorporating fundamental insights from previous solutions while keeping the focus on standardization and benchmarking.

\section{RL baselines}
\label{suppl:rl_baselines}

In this section, we briefly introduce the baseline RL algorithms employed in our evaluation, referring to the original papers for exhaustive details about these methods \citep{dqn, ppo, sac, td3}.

\textbf{DQN} \citep{dqn}. A DQN agent uses a neural network to approximate the action value function $Q$ by taking as input the state of the environment and outputting $Q$-values for every possible action. During training, the agent uses an $\epsilon$-greedy policy to select random actions or follow the greedy policy on these $Q$-values, according to a linearly decaying probability $\epsilon$. The $Q$ network is thus updated to minimize the difference between predicted $Q$-values and a target derived from actual rewards and future $Q$-values. To deal with overestimation, we use Double-DQN \citep{doubledqn} and decouple action selection from action evaluation using a target $Q$ network. Due to its value-based nature, a DQN agent can only consider discrete (topological) actions.

% directly approximates a policy by learning its parameters using a computationally tractable clipped objective. By learning different probability distributions, a PPO agent can deal with continuous (redispatching) and discrete (topological) actions. The lagrangian version is used in the constrained topological environments as it represents a popular and good performing baseline for CMDP-based tasks \citep{ijcai2021constraints}.

\textbf{PPO} \citep{ppo} and \textbf{Lagrangian PPO} (LagrPPO) \citep{LagrangianPPO}. A PPO agent uses its neural network to directly approximate a policy. The agent learns the policy parameters by simplifying the TRPO \citep{trpo} algorithm, using a computationally tractable clipped objective. This clipping mechanism prevents large changes to the policy that could destabilize the training. At a high level, such a surrogate objective balances policy improvement and limits the divergence between policy updates. To drive the policy training, PPO also learns an advantage function to determine how much better (or worse) taking an action is compared to the expected value. By employing different probability distributions as a policy, a PPO agent can deal with both continuous (redispatching) and discrete (topological) actions. The Lagrangian version applies the same intuitions while learning additional value functions for each constraint. It then changes to policy training by considering the Lagrangian discussed in Section \ref{sec:preliminaries}. In more detail, Lagrangian algorithms take gradient ascent steps in $\pi$ and descent steps in $\bm{\lambda}$ to trade off safety and task performance. These methods focus on satisfying the constraints using penalties $\bm{\lambda}$ that grows unbounded when constraints are violated. When constraints are satisfied, $\bm{\lambda}$ scale down (to zero), allowing the algorithm to maximize the task objective. 

\textbf{SAC} \citep{sac}. Similarly to PPO, a SAC agent learns different networks to maintain a policy and two value functions that mitigate positive bias in value estimates. Overall, the agent maximizes both the expected return and the entropy of the policy. The entropy term encourages exploration by promoting stochastic policies, which helps prevent premature convergence to suboptimal policies. In terms of actions, the SAC agent can deal with the same action types as PPO.

\textbf{TD3} \citep{td3}. A TD3 agent learns multiple networks similarly to SAC. However, unlike the stochastic policies learned by PPO and SAC, TD3 learns a deterministic policy and can only deal with continuous actions. To encourage exploration, the agent does not maximize the entropy of the policy but adds noise to the output of the policy network.

\section{Environments}
\label{suppl:env}
\begin{table}[b]
\centering
\vspace{-10pt}
\caption{Action space sizes for the considered environments. Left: Difficulty for environments with a (discrete) topology-based action space. Right: (continuous) redispatching and curtailment tasks.}
\label{tab:action_spaces}
\setlength{\tabcolsep}{3pt}
\begin{tabular}{lcccccc}
\toprule
                       & \multicolumn{6}{c}{\textit{\textbf{\# Actions per difficulty level}}}                                                                                    \\ \midrule
                       & \multicolumn{5}{c}{\textbf{Topology (T)}}                                                                   & \textbf{Redispatching and curtailment (R)} \\ \midrule
\textit{\textbf{}}     & \textit{\textbf{0}} & \textit{\textbf{1}} & \textit{\textbf{2}} & \textit{\textbf{3}} & \textit{\textbf{4}} & \textit{\textbf{0}}                        \\ \midrule
\textbf{bus14}         & 50                  & 209                 & -                   & -                   & -                   & 6                                          \\
\textbf{bus36-M}       & 50                  & 302                 & 1829                & 11071               & 66978               & 22                                         \\
\textbf{bus36-MO-v0}   & 50                  & 302                 & 1829                & 11071               & 66978               & 22                                         \\
\textbf{bus36-MO-v1}   & 50                  & 302                 & 1829                & 11071               & 66978               & 22                                         \\
\textbf{bus118-M}      & 50                  & 308                 & 1903                & 11744               & 72461               & 69                                         \\
\textbf{bus118-MOB-v0} & 50                  & 309                 & 1914                & 11849               & 73328               & 69                                         \\
\textbf{bus118-MOB-v1} & 50                  & 309                 & 1915                & 11852               & 73357               & 69                                         \\ \bottomrule
\end{tabular}
\end{table}

As discussed in Section~\ref{sec:formalization}, here we introduce the different levels of difficulty for the topological-based environments, as well as the reward function employed in all the tasks.
Each increasing level of task difficulty corresponds to a higher dimensional discrete action space. Table~\ref{tab:action_spaces} summarizes the difficulty levels and the corresponding total number of actions. 

\subsection{Action Spaces Analysis} 
\label{suppl:action_space}

In this section, we visually analyze the action spaces of one representative environment for each power grid size (i.e., bus14, bus36-MO-v0, bus118-M).

\begin{figure}[h]
    \centering
    \includegraphics[width=0.6\textwidth]{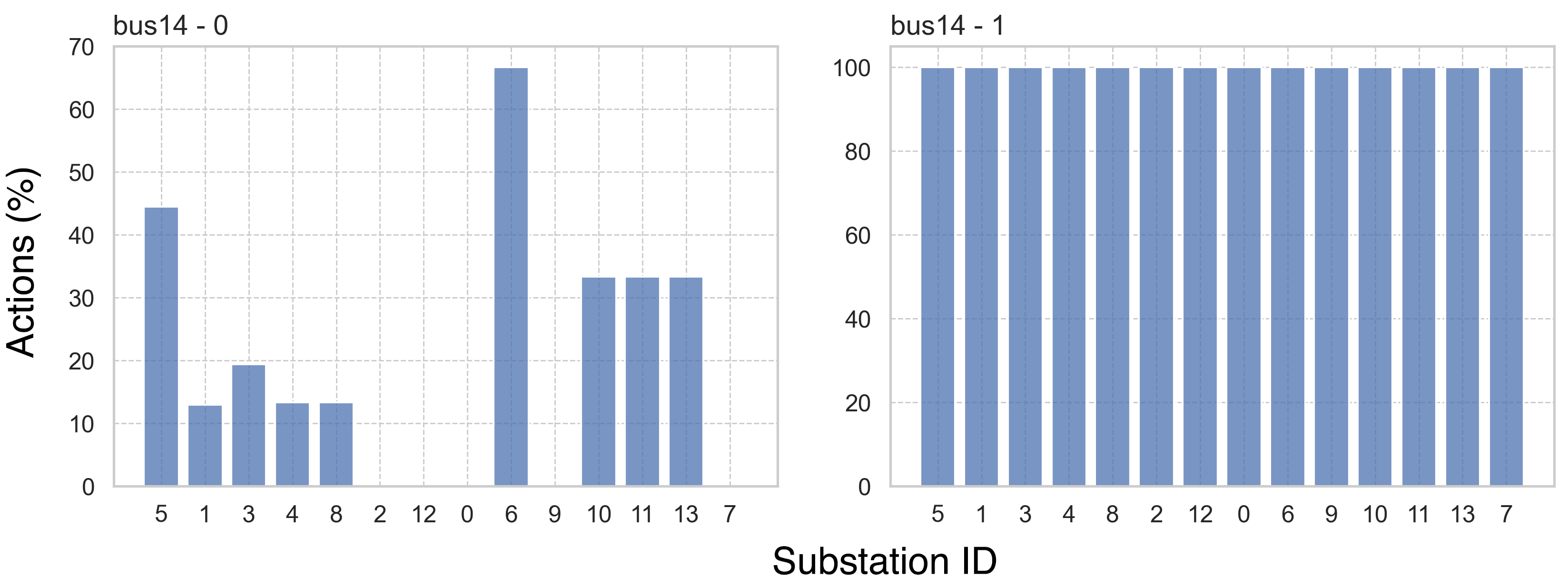}
    \caption{Percentage of actions considered for each substation within the action space for bus14 (discrete) topological tasks (difficulty level is indicated with the number on the top left).}
    \label{fig:actions_rate_bus14}
    \vspace{-5pt}
\end{figure}

\begin{figure}[h]
    \centering
    \includegraphics[width=1.0\textwidth]{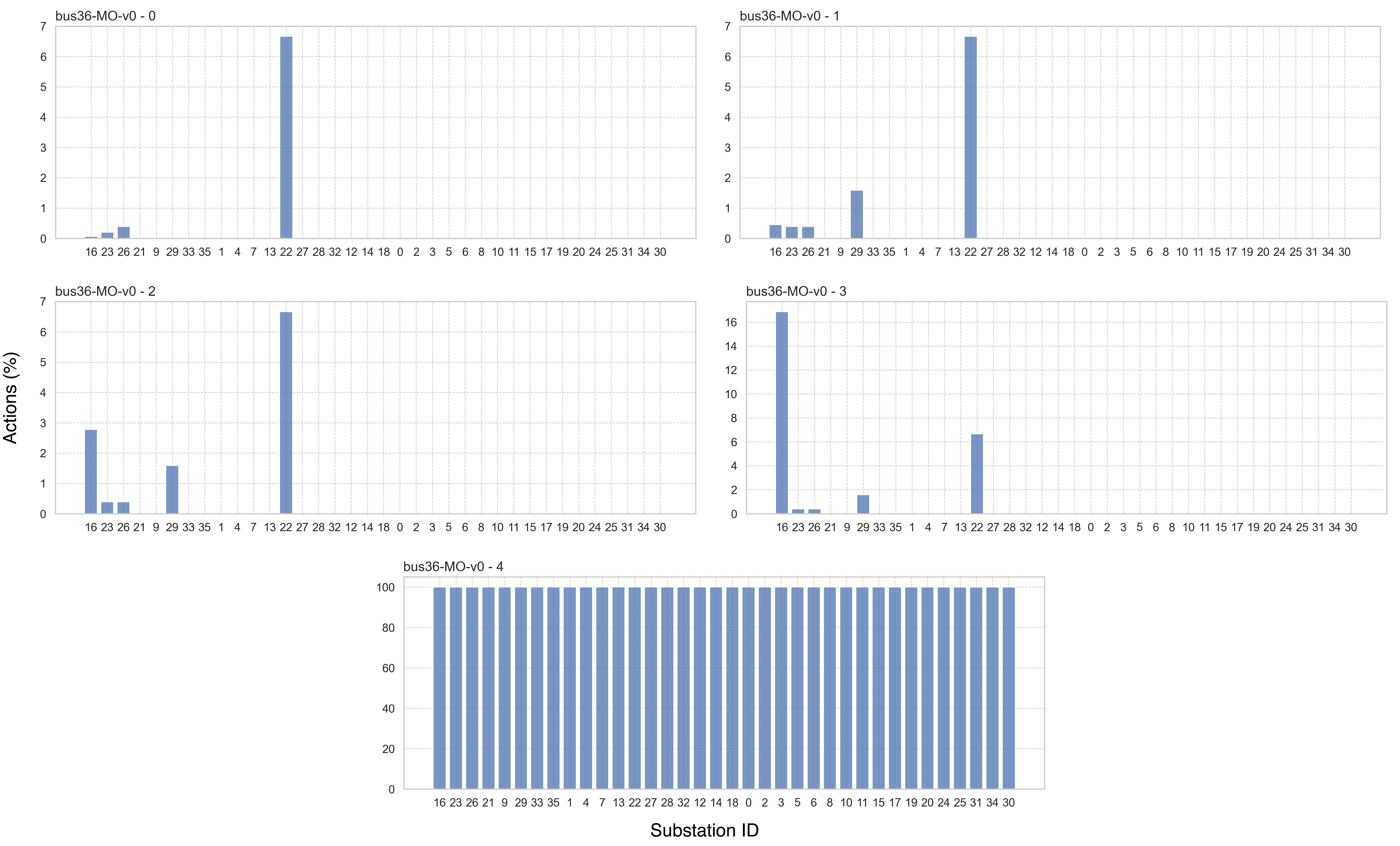}
    \caption{Percentage of actions considered for each substation within the action space for bus36-MO-v0 (discrete) topological tasks (difficulty level is indicated with the number on the top left).}
    \label{fig:actions_rate_bus36}
    \vspace{-5pt}
\end{figure}

\begin{figure}[h]
    \centering
    \includegraphics[width=0.9\textwidth]{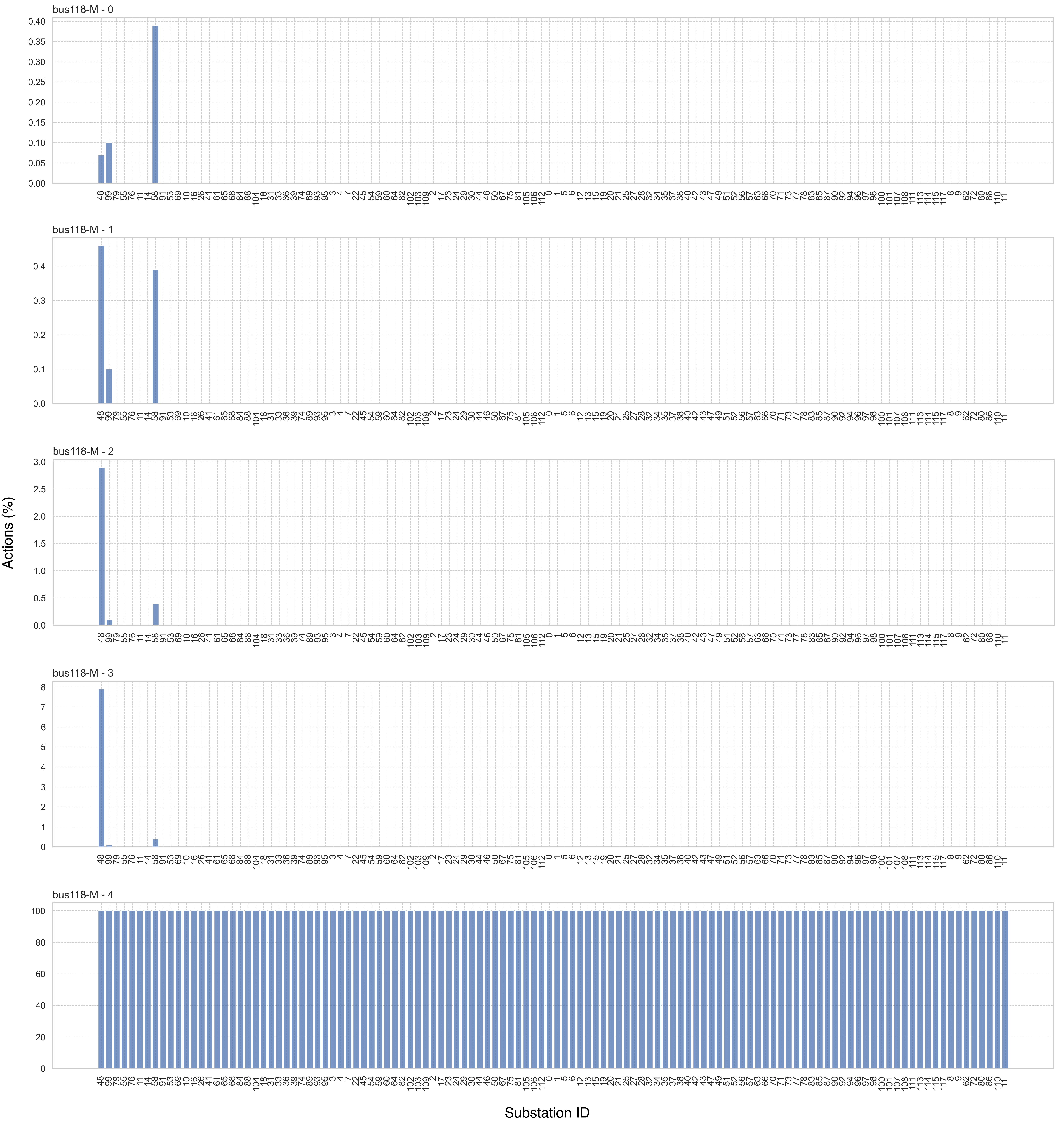}
    \caption{Percentage of actions considered for each substation within the action space for bus118-M (discrete) topological tasks (difficulty level is indicated with the number on the top left).}
    \label{fig:actions_rate_bus118}
    \vspace{-5pt}
\end{figure}

For each difficulty level, \Cref{fig:actions_rate_bus14,,fig:actions_rate_bus36,,fig:actions_rate_bus118} show the percentage of actions considered for each substation within the action space. The x-axis lists the substation IDs in descending order based on the number of available actions. The y-axis represents the ratio of actions used in the action space to the total number of available actions for each substation. Consequently, the highest difficulty level indicates that the action space includes all possible actions for all substations. Overall, this analysis suggests that the substation with the most electric components (i.e., the most possible topologies) is best suited to handle contingencies.


\begin{figure}[h]
    \centering
    \includegraphics[width=0.35\textwidth]{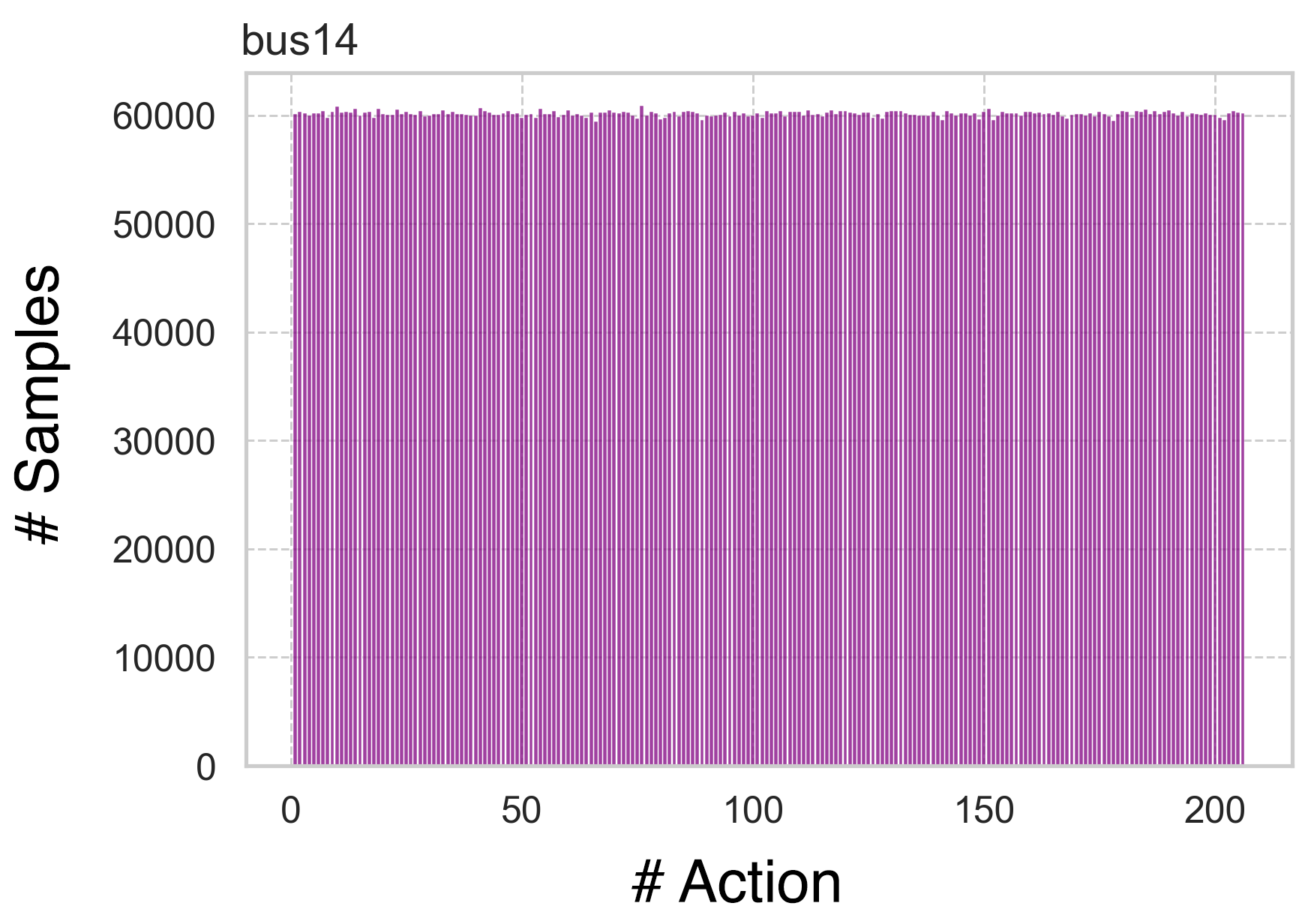} 
    \vspace{-5pt}
    \caption{Number of times each action is sampled over the ranking process time.}
    \label{fig:actions_sample_bus14}
    \vspace{-5pt}
\end{figure}

\begin{figure}[h]
    \centering
    \includegraphics[width=1.0\textwidth]{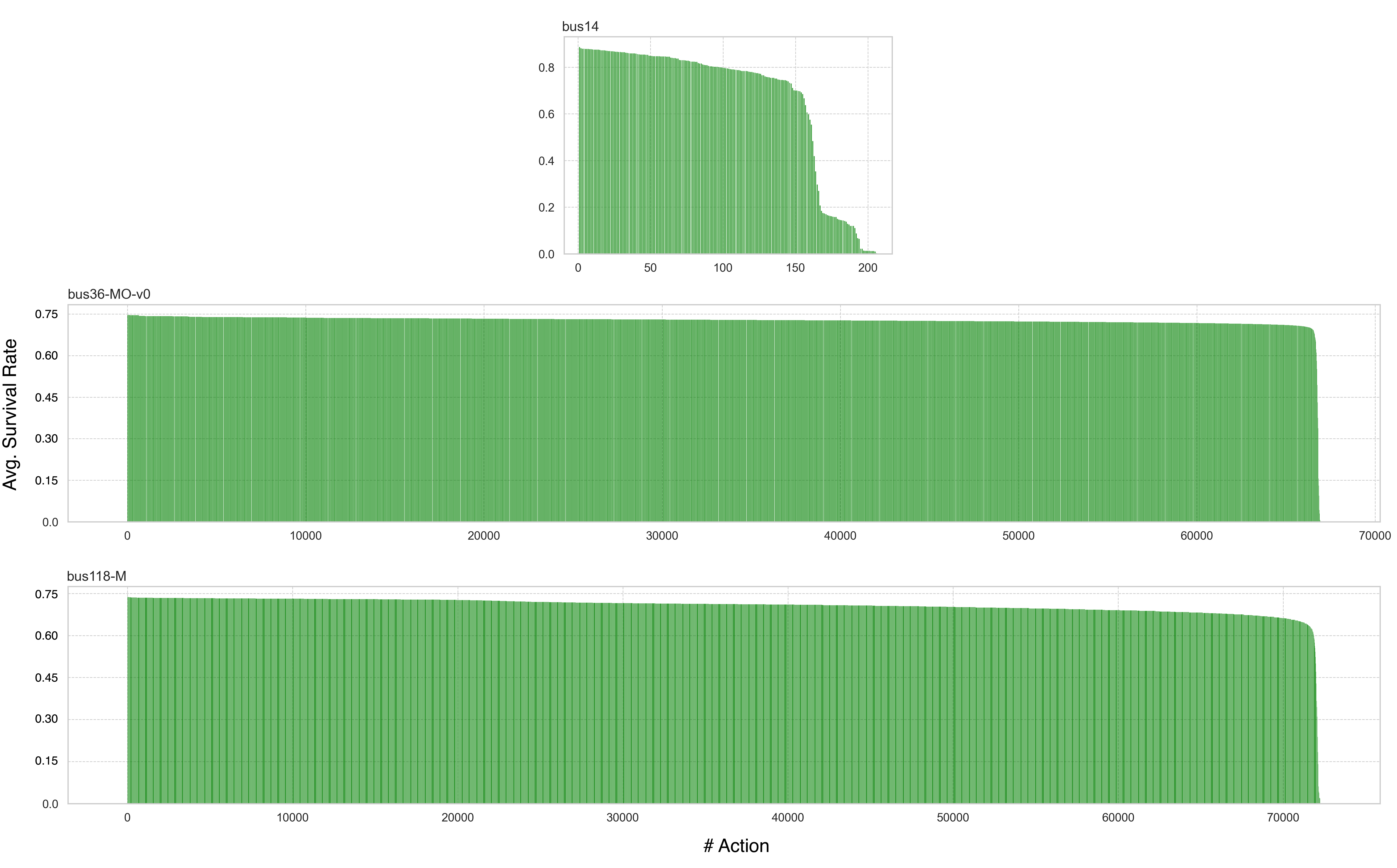}
    \caption{Average survival rate of the action spaces after the ranking process time.}
    \label{fig:actions_ranking}
    \vspace{-5pt}
\end{figure}

\newpage
\Cref{fig:actions_sample_bus14,,fig:actions_ranking} then present the data collected during the action ranking mechanism described in Section \ref{sec:formalization}.

As a sanity check, \Cref{fig:actions_sample_bus14} shows an example of the uniform sampling strategy used to select which action to simulate at each simulation step. The x-axis shows the total number of actions for the bus14 (discrete) topological task; the y-axis indicates the number of times each action was sampled during the ranking process.

\Cref{fig:actions_ranking} shows the final ranking of the actions for the three representative environments. The x-axis shows the total number of actions for each task; the y-axis indicates the average survival rate of each action during the ranking process. Crucially, most of the actions are relevant (i.e., with a high survival rate) in the tasks, motivating the increasing levels of difficulty we proposed for the (discrete) topological environments.

% \subsection{Reward} 
% \label{suppl:reward}

% To promote the survival of the grid, the agent gets a cumulative positive constant $R_{\text{survive}}$ for each step, normalized by the total length of a training episode (normalized $\in [0, 1]$). The capacity reward $R_{\text{capacity}}$ is based on how many lines are used (the lower the better and goes negative in case of overflow) and is set to a fixed penalty value for disconnected lines (normalized $\in [-1, 1]$). The costs component $R_{\text{cost}}$ assigns a cost to redispatching action and penalizes energy losses (normalized $\in [-1, 0]$). Finally, the topology component  $R_{\text{topology}}$ incentivizes the agent to revert to the original topology by computing the distance of the current grid to the one at time $0$ (normalized $\in [0, 1]$). The total reward $R$ an agent gets at each step is then a weighted sum:
% $$
% R = \alpha R_{\text{survive}} + \beta R_{\text{capacity}} + \eta R_{\text{cost}} + 
% \begin{cases}
%     \omega R_{\text{topology}} & \text{if topology actions,} \\
%     0 & \text{otherwise.}
% \end{cases}. 
% $$

\section{State Space}
\label{suppl:state_space}

Regardless of the task, at a certain time-step $t$ an agent gets the following set of features: $[t, \text{Gen}_P, \text{Gen}_{\theta}, \text{Load}_P, \text{Load}_\theta, \rho, \text{Cooldown}_{\text{lines}}]$.
Additionally, based on the nature of the task, the agent can observe additional features as follows:
\begin{itemize}
    \item Topological actions: when an agent operates using (discrete) topological actions, it observes $[$\text{Topo\textsubscript{vect}}, Line\textsubscript{status}, Time\textsubscript{overflow}, Time\textsubscript{sub-cooldown}$]$.
    \item Redispatching actions: when an agent operates using (continuous) redispatching actions, it observes $[$Tg\textsubscript{dispatch}, Curr\textsubscript{dispatch}, Gen\textsubscript{margin-up}, Gen\textsubscript{margin-down}$]$.
    \item Curtailment actions: when an agent operates using (continuous) curtailment actions, it observes $[$Gen\textsubscript{P\textsubscript{curt}}, Curtail, Curtail\textsubscript{limit}$]$.
    \item Maintenance: when the task has maintenance contingencies (see \Cref{tab:environments}), the agent gets $[$Time\textsubscript{next-maint}, Duration\textsubscript{next-maint}$]$.
    \item Storage: when the task has batteries (see \Cref{tab:environments}), the agent gets $[$Storage\textsubscript{charge}, Storage\textsubscript{power\textsubscript{tg}}, Storage\textsubscript{power}, Storage\textsubscript{$\theta$}$]$.
\end{itemize}

Such a distinction is useful to reduce the size of the space the agent can observe when there are features that are not relevant to a specific task. For example, if an agent uses only discrete actions (topology), then everything related to target dispatch, actual dispatch, and storage is irrelevant as they will not change. Likewise, if an agent uses only continuous actions, it is not necessary to include features related to ``topology'' as they will not be modified. Additionally, all the features related to voltage (e.g., voltage for generators, loads, $\dots$) and reactive values (e.g., reactive power for generator, loads, $\dots$) can be neglected. 

For the interested RL practitioner, we refer to the original Grid2Op documentation for exhaustive descriptions of these features \citep{grid2op}.

\section{Environmental Impact}
\label{suppl:env_impact}
Despite each training run being ``relatively'' computationally inexpensive due to the use of CPUs, the experiments of our evaluation led to cumulative environmental impacts due to computations that run on computer clusters for an extended time. Our experiments were conducted using a private infrastructure with a carbon efficiency of $\approx 0.275 \frac{\text{kgCO$_2$eq}}{\text{kWh}}$. Total emissions are estimated to be $\approx216.56 \text{kgCO$_2$eq}$ using the \href{https://mlco2.github.io/impact#compute}{Machine Learning Impact calculator}, and we purchased offsets for this amount through \href{https://www.treedom.net}{Treedom}.
We do not directly estimate or offset other categories of environmental impacts (such as water usage or embodied hardware impacts), though recognizing that these are additionally important to consider.

\section{Hyperparameters}
\label{suppl:hyperparameters}

%The dimensions of the networks are different based on the task (and difficulty). We list these values in \Cref{tab:network_sizes}.
%\input{Tables/net_dimension}

\Cref{tab:hyperparameters} lists the hyperparameters considered during our initial grid search and the final (best-performing) parameters used for our experiments.

\begin{table}[h]
\caption{Details of the grid search used to find the best-performing hyperparameters for each algorithm in the topological (T) and redispatching (R) cases.}
\label{tab:hyperparameters}
\begin{tabular}{llll}
\toprule
\multicolumn{1}{c}{\textbf{Algorithm}} & \multicolumn{1}{c}{\textbf{Parameter}} & \multicolumn{1}{c}{\textbf{Grid search}} & \multicolumn{1}{c}{\textbf{Chosen value (T - R)}} \\ \midrule
\textbf{Shared}                        & \textit{N° parallel envs}              & 10, 20, 50                                       & 50                                                                   \\
\textbf{}                              & \textit{Learning starts}               & 20000                                    & 20000                                                                \\
\textbf{}                              & \textit{Max gradient norm}             & 10, 20, 50                               & 10                                                                   \\
\textbf{}                              & \textit{Discount $\gamma$}             & 0.9, 0.95, 0.99                          & 0.9                                                                  \\
\textbf{}                              & $\alpha$                    & 0.1, 0.5, 1.0                             & 1.0                                                                  \\
\textbf{}                              & $\beta$                    & 0.1, 0.5, 1.0                             & 0.5                                                                  \\
\textbf{}                              & $\eta$                    & 0.1, 0.25, 0.5                             & 0.5                                                                  \\
\textbf{}                              & $\lambda$                    & 0, 50                             & 0 (TLO), 50 (LSI)                                                                 \\ \midrule
\textbf{DQN}                           & \textit{Train frequency}               & 50, 100, 1000                            & 50                                                                   \\
\textbf{}                              & \textit{Target network update}         & 1000, 5000, 10000                         & 5000                                                                 \\
\textbf{}                              & \textit{Buffer size}                   & 100000, 250000, 500000, 1000000          & 1000000                                                               \\
\textbf{}                              & \textit{Batch size}                    & 64, 128, 256                             & 128                                                                  \\ 
\textbf{}                              & \textit{Learning rate}                 & 0.003, 0.0003, 0.00003                   & 0.0003                                                               \\
\textbf{}                              & \textit{$\epsilon$-decay fraction}     & 0.3, 0.5 0.7                             & 0.5                                                                  \\ \midrule
\textbf{PPO}                           & \textit{N° steps (total)}                      & 10000, 20000, 50000                      & 20000                                                                \\
\textbf{}                              & \textit{N° minibatches}                    & 4, 8, 12                             & 4                                                                  \\ 
\textbf{}                              & \textit{N° update epochs}              & 20, 40, 80                               & 40                                                                   \\
\textbf{}                              & \textit{Actor learning rate}           & 0.003, 0.0003, 0.00003                   & 0.0003 - 0.00003                                                     \\
\textbf{}                              & \textit{Critic learning rate}          & 0.003, 0.0003, 0.00003                   & 0.0003 - 0.00003                                                     \\
\textbf{}                              & \textit{$\epsilon$-clip}               & 0.1, 0.2, 0.3                            & 0.2                                                                  \\ \midrule
\textbf{SAC}                           & \textit{Train frequency}               & 50, 100, 1000                            & 50                                                                   \\
\textbf{}                              & \textit{Actor delayed update}          & 2, 4                                     & 2                                                                    \\
\textbf{}                              & \textit{Noise clip}                    & 0.5                                      & 0.5                                                                  \\
                                       & \textit{Buffer size}                   & 100000, 250000, 500000, 1000000          & 500000                                                               \\
                                       \textbf{}                              & \textit{Batch size}                    & 64, 128, 256                             & 128                                                                  \\ 
                                       & \textit{Actor learning rate}           & 0.003, 0.0003, 0.00003                   & 0.0003 - 0.00003                                                     \\
                                       & \textit{Critic learning rate}          & 0.003, 0.0003, 0.0003                    & 0.0003 - 0.00003                                                     \\
                                       & \textit{Entropy regularization}        & 0.02, 0.2, 0.4                                      & 0.2                                                                  \\\midrule
\textbf{TD3}                           & \textit{Actor delayed update}          & 2, 4                                     & 2                                                                    \\
                                       & \textit{Buffer size}                   & 100000, 250000, 500000, 1000000          & 1000000                                                               \\
                                       \textbf{}                              & \textit{Batch size}                    & 64, 128, 256                             & 128                                                                  \\ 
                                       & \textit{Actor learning rate}           & 0.003, 0.0003, 0.00003                   & 0.0003 - 0.00003                                                     \\
                                       & \textit{Critic learning rate}          & 0.003, 0.0003, 0.00003                   & 0.0003 - 0.00003                                                     \\
                                       & \textit{$\tau$}                        & 0.005, 0.0005                            & 0.005                                                                \\
                                       & \textit{Policy noise}                  & 0.2                                      & 0.2                                                                  \\
                                       & \textit{Exploration noise}             & 0.1                                      & 0.1                                                                  \\ \bottomrule
\end{tabular}
\end{table}

\clearpage
\newpage
\section{Omitted Figures in Section 5}
\label{suppl:results}
\begin{figure}[h]
    \centering
    \includegraphics[width=1.0\textwidth]{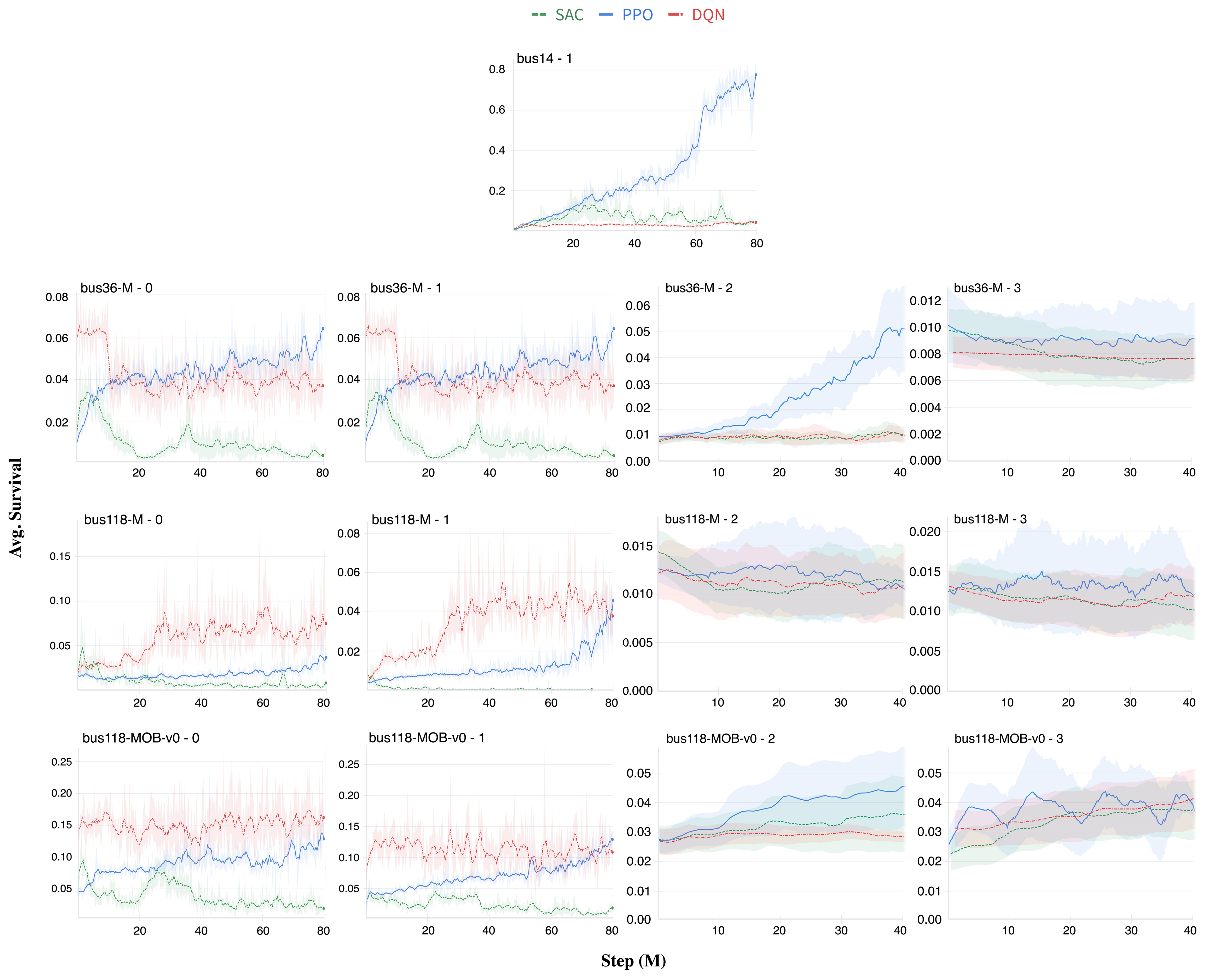}
    \caption{Average survival rate for the discrete topological case in \textit{bus14, bus36-M, bus118-M, bus118-MOB-v0} using the SAC, PPO, and DQN baselines. We indicate the difficulty level (ranging from 0 to 3) next to the environment identifier.}
    \label{fig:topo_result}
    \vspace{-5pt}
\end{figure}

The following figures report plots collected on Wandb \citep{wandb}. Figure \ref{fig:topo_result} shows the training curves for the remaining (discrete) topological action spaces. Due to the strict time limit imposed on the computation nodes (see Section 4) and the different computational requirements of the algorithms, not all the baselines perform the same number of steps in the time limit and the experiments with $36$ and $118$ bus consider 5 runs. The demands and limited performance of the topological baselines led us to exclude the results with the complete action space (i.e., difficulty set to 4). Additionally, despite the grid search of Table \ref{tab:hyperparameters}, some baselines achieved lower performance than expected (e.g., SAC and DQN in the \textit{bus14} scenarios). % We will keep working on the benchmark to find better parameters, run longer experiments, and keep the following figures updated.

\begin{figure}[h]
    \centering
    \includegraphics[width=1.0\textwidth]{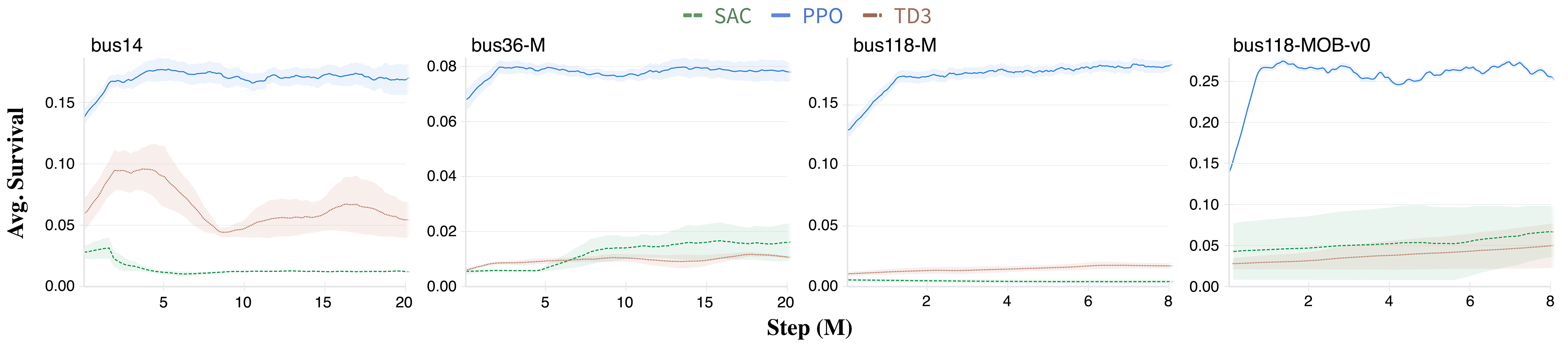}
    \caption{Average survival rate for the continuous redispatching case in \textit{bus14, bus36-M, bus118-M, bus118-MOB-v0} using the SAC, PPO, and TD3 baselines.}
    \label{fig:redisp_result}
    \vspace{-5pt}
\end{figure}

\newpage
Figure \ref{fig:redisp_result} shows the training curves for the (continuous) redispatching action spaces.

\begin{figure}[h]
    \centering
    \includegraphics[width=0.8\textwidth]{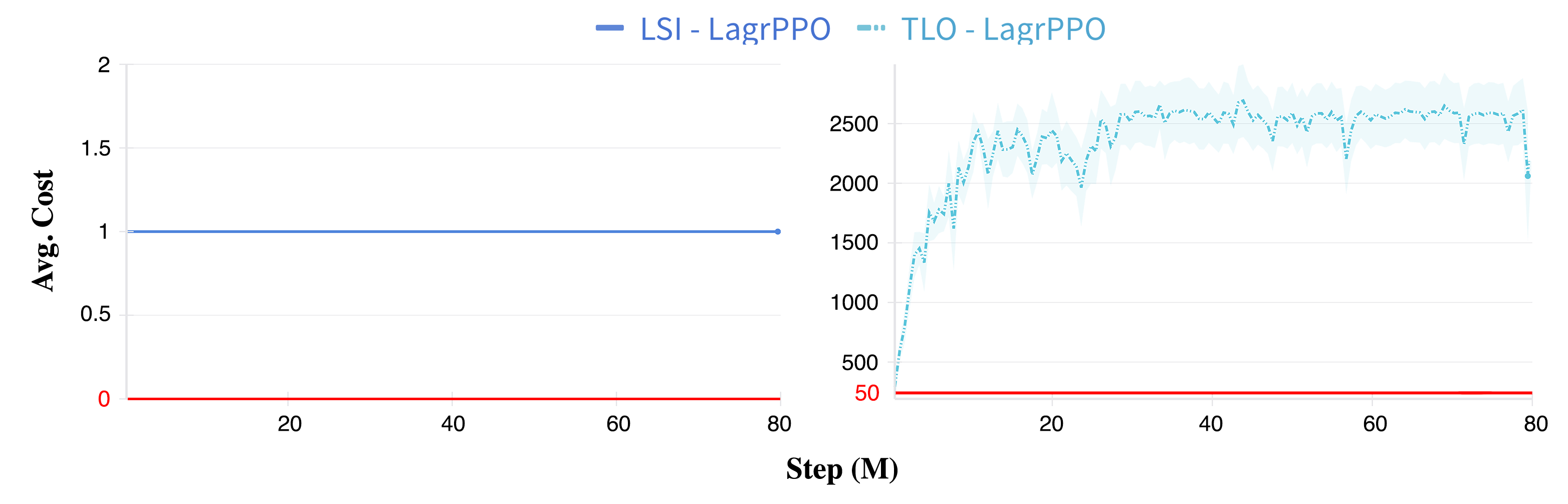}
    \caption{Average cost rate for constrained case in the representative \textit{bus14} using the LagrPPO baseline. The constrained threshold (red line) is set to 0 and 50 for the TLO and LSI cases, respectively.}
    \label{fig:cost_result}
    \vspace{-5pt}
\end{figure}

Figure \ref{fig:cost_result} shows the cost obtained over the training for the constrained experiments.

\section{Omitted Performance Analysis in Section 5}
\label{suppl:performance analysis}

\begin{figure}[h]
    \centering
    \includegraphics[width=1.0\textwidth]{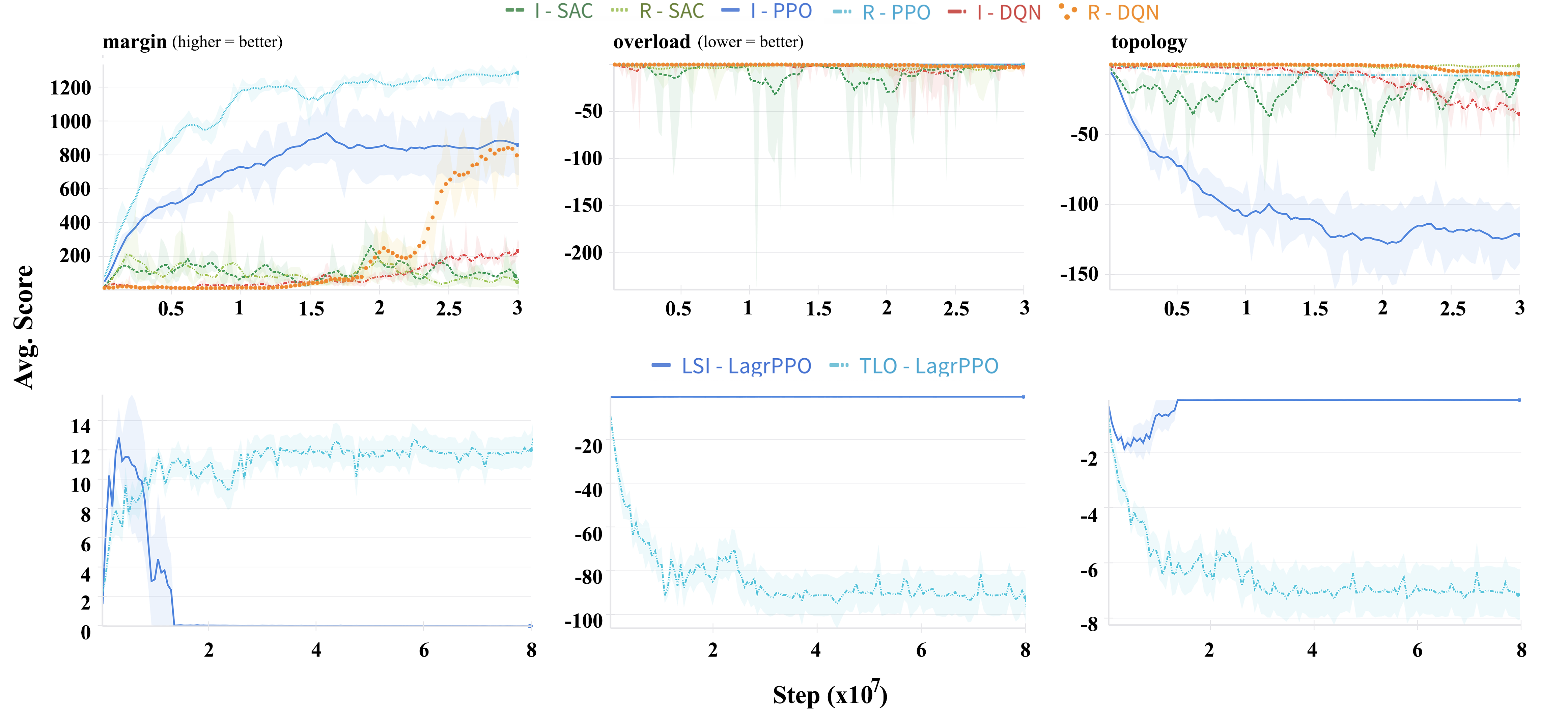}
    \vspace{-10pt}
    \caption{Average score for key operational components across heuristic and constraints-based baselines (reported on separate rows). The first column represents line margins, where higher values indicate better contingency management. The second column tracks overload penalties, with lower values reflecting improved grid stability. The third column captures topology modifications, showing the extent to which agents reconfigure the grid.}
    \label{fig:bus14_paper_analysis}
    \vspace{-5pt}
\end{figure}

\end{document}